\documentclass[lettersize,journal]{IEEEtran}
\usepackage{amsmath,amssymb,amsfonts}
\usepackage{algorithmicx,algorithm}
\usepackage{amsthm,amsmath,amssymb}
\usepackage{mathrsfs}
\usepackage{array}
\usepackage{textcomp}
\usepackage{stfloats}
\usepackage{verbatim}
\usepackage{graphics} % for pdf, bitmapped graphics files
\usepackage{epsfig} % for postscript graphics files
\usepackage{times} % assumes new font selection scheme installed
\usepackage{amsmath} % assumes amsmath package installed
\usepackage{amssymb}  % assumes amsmath package installed
\usepackage{cite} % 导入引用的包，能够使用\cite
\usepackage{graphicx}
\usepackage{subfigure} %子图
\usepackage{picinpar}
\usepackage{url}
\usepackage{flushend}
\usepackage{colortbl}
\usepackage{soul}
\usepackage{multirow}
\usepackage{pifont}
\usepackage{color}
\usepackage{alltt}
\usepackage{hyperref}
\usepackage{float}%需要把图片放在当前位置，不容改变，可以用float宏包的[H]选项
\usepackage{enumerate}
\usepackage{siunitx}
\usepackage{breakurl}
\usepackage{epstopdf}
\usepackage{pbox}
\usepackage{booktabs}
\usepackage{makecell} %表格换行
\usepackage{threeparttable} %加脚注
\usepackage{balance}%最后一页 参考文献不平衡
\usepackage[justification=centering,labelsep=period]{caption} %全局图片标题居中
\usepackage{color} %颜色
\usepackage{gensymb} %度
\hypersetup{colorlinks=true, linkcolor=black} %修改链接颜色

\usepackage{adjustbox}%用于highlight
\usepackage{xcolor}%用于highlight
\usepackage{soul}
\usepackage{pdfpages}
\soulregister\cite7 % 针对\cite命令
\soulregister\citep7 % 针对\citep命令
\soulregister\citet7 % 针对\citet命令
\soulregister\ref7 % 针对\ref命令
\soulregister\pageref7 % 针对\pageref命令

\usepackage{bm}
\usepackage{geometry}
\title{\LARGE \bf
ESVIO: Event-based Stereo Visual Inertial Odometry 
}

\author{Peiyu Chen$^{*}$,   Weipeng Guan$^{*}$,   Peng Lu
\thanks{
    * Equal contribution.
    
    The authors are with the Adaptive Robotic Controls Lab (ArcLab), Department of Mechanical Engineering, The University of Hong Kong, Hong Kong SAR, China. \url{lupeng@hku.hk}. 
    
    This research is supported by General Research Fund under Grant No. 17204222, the Seed Funding for Strategic Interdisciplinary Research Scheme and Platform Technology Fund.
}%
}

\begin{document}   %正式开始这个文档
\maketitle  %不要加页面，避免编译不过
\thispagestyle{headings} 
\pagestyle{headings} %添加这个让每页都有   % empty - 没有页眉和页脚  % plain - 没有页眉，页脚包含一个居中的页码  % headings - 没有页脚，页眉包含章/节或者字节的名字和页码 % myheadings - 没有页脚，页眉包含有页码

%%%%%%%%%%%%%%%%%%%%%%%%%%%%%%%%%%%%%%%%%%%%%%%%%%%%%%%%%%%%%%%%%%%%%%%%%%%%%%%%
\begin{abstract}
% Event cameras asynchronously output low-latency event streams based on brightness change without motion blur, which provides great opportunities for state estimation under aggressive motion or broad dynamic range scenes.
% 
Event cameras that asynchronously output low-latency event streams provide great opportunities for state estimation under challenging situations.
Despite event-based visual odometry having been extensively studied in recent years, most of them are based on the monocular, while few research on stereo event vision.
In this paper, we present ESVIO, the first event-based stereo visual-inertial odometry, which leverages the complementary advantages of event streams, standard images, and inertial measurements.
Our proposed pipeline includes the ESIO (purely event-based) and ESVIO (event with image-aided), which achieves spatial and temporal associations between consecutive stereo event streams. 
A well-design back-end tightly-coupled fused the multi-sensor measurement to obtain robust state estimation.
% In addition, the motion compensation method is designed to emphasize the edge of scenes by warping each event to reference moments with IMU and \hl{the state from back-end}.
We validate that both ESIO and ESVIO have superior performance compared with other image-based and event-based baseline methods on public and self-collected datasets.
Furthermore, we use our pipeline to perform onboard quadrotor flights under low-light environments.
Autonomous driving data sequences and real-world large-scale experiments are also conducted to demonstrate long-term effectiveness.
We highlight that this work is a real-time, accurate system that is aimed at robust state estimation under challenging environments.

\end{abstract}

%%%%%%%%%%%%%%%%%%%%%%%%%%%%%%%%%%%%%%%%%%%%%%%%%%%%%%%%%%%%%%%%%%%%%%%%%%%%%%%%
\begin{IEEEkeywords}
Visual-Inertial SLAM, sensor fusion, aerial systems: perception and autonomy. 
\end{IEEEkeywords}

%%%%%%%%%%%%%%%%%%%%%%%%%%%%%%%%%%%%%%%%%%%%%%%%%%%%%%%%%%%%%%%%%%%%%%%%%%%%%%%%
\vspace{-1.0em}%调整与上文的距离
\section*{MULTIMEDIA MATERIAL}  %不带序号
\label{MULTIMEDIA MATERIAL}
{\footnotesize \textbf{Website}: \url{https://github.com/arclab-hku/Event_based_VO-VIO-SLAM}.}

% {\footnotesize \textbf{Youtube}: \url{https://youtu.be/XqAm1q0alNY}.}

{\footnotesize \textbf{Video Demo}: \url{https://b23.tv/V23SVzC}.}

{\footnotesize \textbf{Code Link}: \url{https://github.com/arclab-hku/ESVIO}.}

% {\footnotesize \textbf{Supplementary Material}: \url{https://github.com/arclab-hku/Event_based_VO-VIO-SLAM/tree/main/ESVIO/supply}.}

%%%%%%%%%%%%%%%%%%%%%%%%%%%%%%%%%%%%%%%%%%%%%%%%%%%%%%%%%%%%%%%%%%%%%%%%%%%%%%%%
\section{INTRODUCTION}
\label{INTRODUCTION}

\IEEEPARstart
% {S}{imultaneous} Localization and Mapping (SLAM), Visual Odometry (VO), and Visual-Inertial Odometry(VIO) based on standard cameras have been extensively researched during the last two decades\cite{CPYHKU:ORB-SLAM3}\cite{CPYHKU:VINS-MONO}.
% However, due to the inherent limitations of standard cameras (motion blur and low dynamic range), VO/VIO/SLAM based on standard cameras still struggle to cope with extreme situations, e.g. aggressive motion and difficult light conditions.
{E}{vent} cameras are novel bio-inspired sensors\cite{CPYHKU:EVENT-SURVEY}, which have a high dynamic range (140 dB compared to 60 dB of standard cameras) to handle broad illumination conditions.
%%%%%%%%%%%%%%%%%%***********************************************************************************************************************************************************%%%%%%%%%%%%%%%%%
\begin{figure}[htb]  %%(h 此处（here） t 页顶（top）b 页底（bottom） p 独立一页（page）)
        % \setlength{\abovecaptionskip}{-1.0em}%调整图片标题与图距离
        % \vspace{-1.0em}%调整表格与正文的距离
        \centering
        \captionsetup{justification=justified}%图题对齐
        \includegraphics[width=1.0\columnwidth]{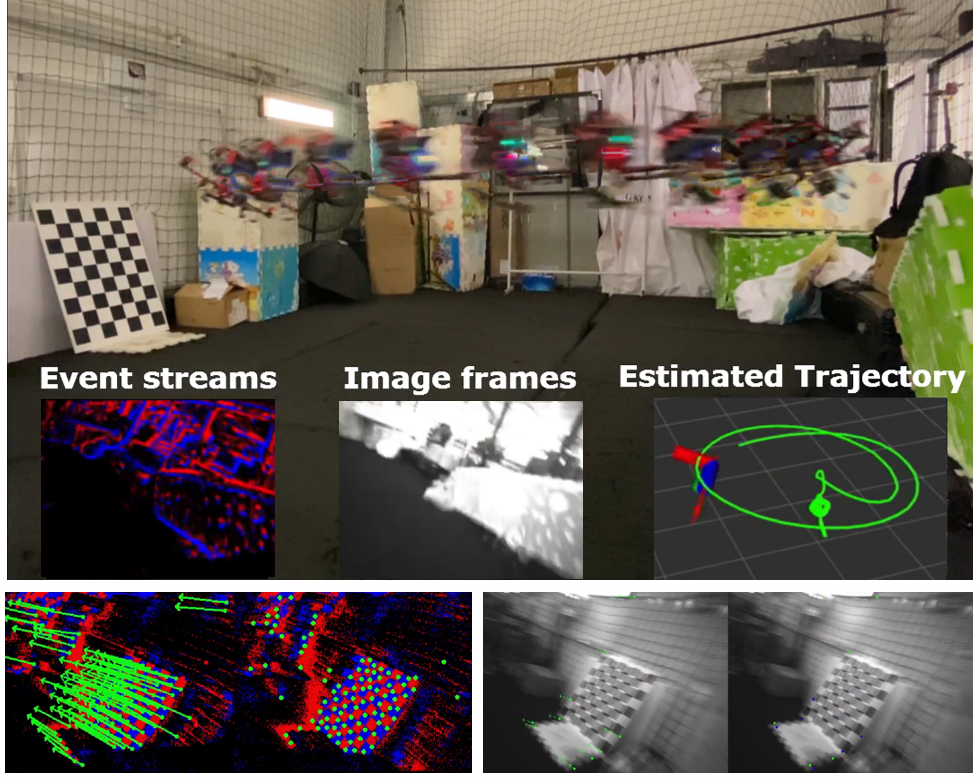}
        \caption{Our ESVIO provides robust and accurate, real-time pose feedback for drones under aggressive motion. 
        Events provide rich and reliable features, while only a few features are tracked in image frames in high-speed motion. 
        \textbf{Left bottom:} stereo event-based feature tracking. 
        \textbf{Right bottom:} stereo image-based feature tracking. }  %设置图片的一个编号以及为图片添加标题
        \label{firstpic}
        \vspace{-2.0em}%调整表格与正文的距离
\end{figure}%
%%%%%%%%%%%%%%%%%%***********************************************************************************************************************************************************%%%%%%%%%%%%%%%%%%
% Unlike standard cameras that output fixed-rate image frames, event cameras have a different paradigm of sensing that responds to pixel-level intensity changes and output asynchronous event streams at low latency.
% This addresses the limitation of standard cameras including motion blur and low dynamic range, as they can tackle high-speed motion without motion blur.
Unlike standard cameras that output fixed-rate image frames, event cameras respond to pixel-level intensity changes and output asynchronous event streams at the latency of microsecond level, which endows these novel sensors to tackle high-speed motion without motion blur.

Most of the existing event-based visual odometers (VO) use monocular event camera\cite{CPYHKU:EVO,CPYHKU:Ultimate-SLAM,CPYHKU:PL-EVIO}, while few research on visual odometry based on stereo event cameras\cite{CPYHKU:ESVO}\cite{CPYHKU:Feature-based-ESVO}. 
Since event cameras output asynchronous event streams rather than fixed-rate image frames, the traditional image-based instantaneous matching cannot be directly implemented on event streams.
For consecutive stereo event streams, merely relying on temporal constraints to extract and match event-corner features might lead to many false correspondences.
Temporal deviations between event streams, noise impacts, different contrast sensitivity of sensors, etc. cause the above problem.
Therefore, it is crucial to extract apposite event-corner features and design proper constraints to achieve data association between stereo event-corner features.

Compared with standard cameras, event cameras do not suffer from motion blur under aggressive motion.
However, when the relative motion between event cameras and scenes is restricted, e.g. in the stationary state, event streams might not be reliably generated and transmitted, whereas standard cameras are able to provide rich information most of the time (e.g. low-speed motion and well-lit scenes). 
Observing this complementarity, leveraging both of the advantages of the aforementioned different sensors in combination with an inertial measurement unit (IMU) results in a robust and accurate visual-inertial odometry (VIO) pipeline\cite{CPYHKU:Ultimate-SLAM} \cite{CPYHKU:PL-EVIO}.

% 做双目事件的意义:单目视觉需要有足够的视差才能恢复出新增角点特征的逆深度，而双目视觉可以直接通过左右相机匹配特征点的极线几何得到特征点的深度信息。
% Monocular vision requires sufficient parallax to recover the inverse depth of the added corner features, while stereo vision can obtain the depth of corner features directly by the epipolar geometry of left and right event cameras.

% 下面这三段讲的是做双目的意义，而这个其实不是本文的重点，我们的重点应该侧重于为什么做双目事件，做双目事件存在什么难点。这三段话，浓缩为2句话。
% Since the inverse depth is not directly available for monocular vision, monocular VO systems usually have to wait until corner features have been tracked for a few frames and the monocular visual has sufficient parallax to recover the inverse depth of the added corner features using triangulation, which also called as the delayed triangulation.
% However, if the parallax observed by the monocular VO is limited, e.g., in-situ rotation, then monocular vision cannot accurately estimate the inverse depth, which might lead to situations such as incorrect scale or even tracking failure\cite{CPYHKU:ORB-SLAM2}.
% Contrastly, stereo vision can recover the inverse depth of corner features by triangulating corner features between the left and right cameras at the current timestamp, which allows the stereo VIO system to be more robust and accurate in special situations compared to monocular vision.

In this paper, we propose, to the best of our knowledge, the first published event-based stereo visual-inertial odometry (ESVIO). Our contributions are summarized as follows:
\begin{enumerate}
\item 
In order to achieve robust state estimation under aggressive motion and low-light scenarios, we propose the first purely event-based stereo VIO (ESIO) pipeline with sliding windows graph-based optimization, and further extend it with image-aided (ESVIO) which tightly integrates stereo event streams, stereo image frames, and IMU together. 

\item
To tackle the problem of event-based stereo feature tracking and matching, we design geometry-based spatial and temporal data associations in consecutive stereo event streams. 
The spatial and temporal constraints ensure accurate and reliable state estimation.
Moreover, a motion compensation method is designed to emphasize the edge of scenes by warping each event.
% to reference moments with IMU and the state from back-end

\item
We evaluate that our ESVIO can achieve state-of-the-art performance on publicly available datasets.
We also release a very challenging event-based VO/VIO dataset, featuring aggressive motion and HDR scenarios. 
Finally, we perform onboard closed-loop quadrotor flight using our ESVIO as the estimator.

\end{enumerate}

The remainder of the paper is organized as follows:
Section \ref{Related Works} introduces the related works. 
Section \ref{Methodology} introduces the methodology of our methods.
Section \ref{Evaluation} presents the experiments and results.
Finally, the conclusion is given in Section \ref{CONCLUSIONS}.

%%%%%%%%%%%%%%%%%%%%%%%%%%%%%%%%%%%%%%%%%%%%%%%%%%%%%%%%%%%%%%%%%%%%%%%%%%%%%%%%
\section{Related Works}
\label{Related Works}
% Visual odometry based on standard cameras has been extensively studied over the past decades. With the commercialization of event cameras, event-based visual odometry has been intensively researched in recent years. Here we review the event-based VO/VIO/SLAM. Ref.\cite{CPYHKU:EVENT-SURVEY} provides a more extensive survey.

\subsection{Event-based Monocular Visual Odometry}
Event-based monocular VO has been intensively researched for challenging scenarios in recent years.
\cite{CPYHKU:kueng2016low} is the first work using feature tracks to achieve event-based VO, which detects features firstly from grayscale frames and then uses event streams tracked features asynchronously. 
EVO\cite{CPYHKU:EVO} proposed a monocular event-based parallel tracking-and-mapping philosophy which applies the image-to-model alignment for tracking and Event-based Multi-View Stereo (EMVS)\cite{CPYHKU:EMVS} for mapping. 
\cite{CPYHKU:Event-based-visual-inertial-odometry} proposed the first event-based VIO that tackles the incomplete estimation of scale and provides accurate 6-DoF state estimation based on Extended Kalman Filter (EKF).
\cite{CPYHKU:ETH-EIO} obtains a discrete number of states based on a spatio-temporal window of event streams, and introduces virtual event frames to achieve nonlinear optimization that refines estimated poses.
Ultimate SLAM\cite{CPYHKU:Ultimate-SLAM} furthered the aforementioned research by combining event streams, image frames, and IMU measurements with nonlinear optimization, which leverages the complementary advantages of event cameras and standard cameras. 
\cite{CPYHKU:Continuous-time-visual-inertial-odometry-for-event-cameras} adopted a continuous-time framework based on cubic spline for smooth trajectory estimation and fused both event streams and IMU together.
DEVO\cite{CPYHKU:DEVO} proposed a novel VO based on a hybrid setup of depth and event cameras, which construct a semi-dense depth map by thresholding time-surface maps.
EKLT-VIO\cite{CPYHKU:EKLT-VIO} integrated an accurate state-of-the-art event-based feature tracker EKLT\cite{CPYHKU:EKLT} with EKF backend to achieve event-based state estimation on Mars-like datasets. 
\cite{CPYHKU:GuanEVIO} proposed a real-time monocular event-based VIO based on graph optimization, which directly utilizes the asynchronous raw events for feature detection.
PL-EVIO\cite{CPYHKU:PL-EVIO} extended the above method to leverage the complementary advantages of standard and event cameras, which tightly combined event-based point features, event-based line features, image-based point features, and IMU measurements together.
% %%%%%%%%%%%%%%%%%%***********************************************************************************************************************************************************%%%%%%%%%%%%%%%%%%
% \begin{figure}[htb]  %%(h 此处（here） t 页顶（top）b 页底（bottom） p 独立一页（page）)
%         % \setlength{\abovecaptionskip}{-1.0em}%调整图片标题与图距离
%         % \vspace{-1.5em}%调整表格与正文的距离
%         \centering
%         \captionsetup{justification=justified}%图题对齐
%         \includegraphics[width=0.85\columnwidth]{image/framework.png}
%         \caption{Framework illustrating the brief pipeline of the proposed ESVIO. The front end starts with preprocessing three different sensor information. The back end tightly fuses instantaneous stereo event residual, temporal stereo event residual, image residual and IMU residual together}  %设置图片的一个编号以及为图片添加标题
%         \label{framework}
%         \vspace{-2.0em}%调整表格与正文的距离
% \end{figure}%
% %%%%%%%%%%%%%%%%%%***********************************************************************************************************************************************************%%%%%%%%%%%%%%%%
\subsection{Event-based Stereo Visual Odometry} 
In contrast to monocular VO, which requires sufficient parallax to recover the depth of corner features, stereo VO can directly obtain the depth of features at the current timestamp.
This solves scale uncertainty and even tracking failure caused by insufficient parallax.
However, most of the recent research on stereo event cameras has focused on depth estimation and constructing semi-dense or dense maps\cite{tulyakov2019learning}\cite{nam2022stereo}, with less research on VO/SLAM fields.
ESVO\cite{CPYHKU:ESVO} proposed the first event-based stereo VO pipeline, which achieves parallel 3D semi-dense mapping thread and tracking thread by maximizing the spatio-temporal consistency of stereo event streams. 
\cite{CPYHKU:Feature-based-ESVO} adopted stereo feature detection and matching with the geometry method, which adopts reprojection error minimization to achieve pose estimation. 
However, these algorithms merely use events to estimate the state, which might result in tracking failure when the system is stationary.
In addition, these aforementioned approaches without combining IMU might lead to losses of visual tracks under textureless areas.
Our work fills a gap based on combining stereo event streams, stereo image frames, and IMU together, which can operate in real-time under high-resolution event streams with better performance than previously proposed methods. 

%%%%%%%%%%%%%%%%%%%%%%%%%%%%%%%%%%%%%%%%%%%%%%%%%%%%%%%%%%%%%%%%%%%%%%%%%%%%%%%%
\section{Methodology}
\label{Methodology}
\begin{figure}[htb]  %%(h 此处（here） t 页顶（top）b 页底（bottom） p 独立一页（page）)
        \vspace{-1.0em}%调整表格与正文的距离
% \begin{adjustbox}{minipage=\linewidth,bgcolor=yellow}        
        \centering
        \captionsetup{justification=justified}%图题对齐
        \includegraphics[width=0.95\columnwidth]{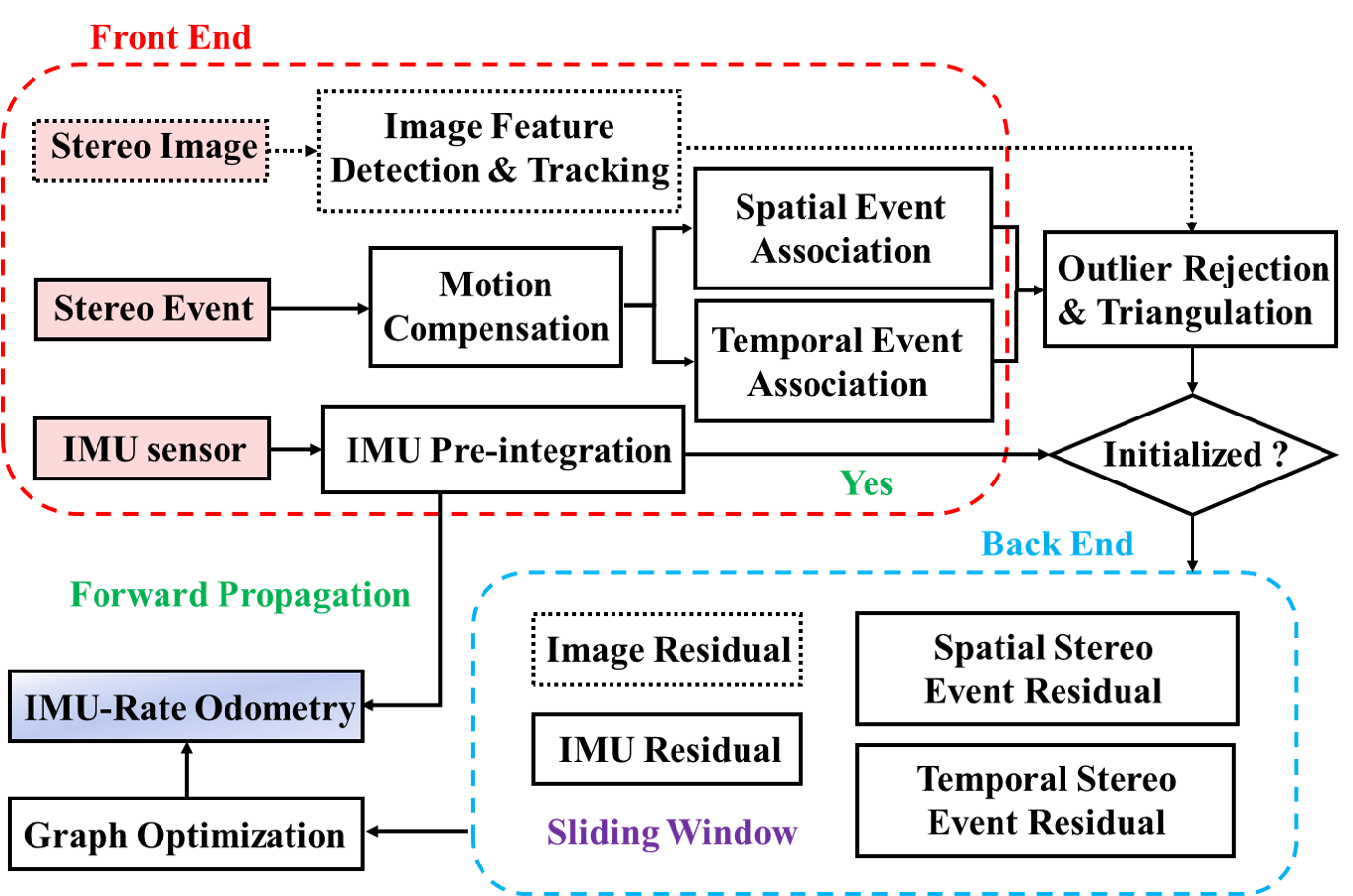}
        \caption{The structure of our ESVIO and ESIO pipeline.}  %设置图片的一个编号以及为图片添加标题
        \label{framework}
% \end{adjustbox}
        \vspace{-0.5em}%调整表格与正文的距离
\end{figure}

% Our proposed ESVIO system takes raw event streams, image frames and inertial measurements as inputs, and manages the residual terms of three different sensors to estimate the pose of the stereo event camera rig.
% A brief framework of our ESVIO is presented in Fig.\ref{framework}. 
% In this paper, we design two frameworks: purely event-based stereo VIO (ESIO) and event with image-aid stereo VIO (ESVIO).
% The pipeline of ESIO can be represented by the ESVIO (shown in Fig.\ref{framework}) without image-based processing. 
Since the procedure of the image measurement is very similar to that of event streams, we only introduce the ESIO in this section.
The core of our framework lies in the pre-processing of raw event streams using motion compensation (Section \ref{Motion Compensation for Event Streams}) and the data association between consecutive stereo event streams in temporal and spatial (Section \ref{Event-based Spatial and Temporal Data Associations}).
After that, we design the event-based constraint for graph optimization (Section \ref{The Construction of Event-based Residual Constraint for the Graph-based Optimization}).
The pipeline of ESIO can be represented by the ESVIO (shown in Fig. \ref{framework}) without image-based processing.

%%%%%%%%%%%%%%%%%%***********************************************************************************************************************************************************%%%%%%%%%%%%%%%%%%
%%%%%%%%%%%%%%%%%%***********************************************************************************************************************************************************%%%%%%%%%%%%%%%%%%

\subsection{Motion Compensation for Event Streams} 
\label{Motion Compensation for Event Streams} 
Motion compensation corrects the curved event streams by aligning events corresponding to the same scene edge.
\cite{CPYHKU:ETH_SCI_Ro_avo} only uses the angular velocity of IMU to achieve rotational compensation while ignoring the effect of translation.
\cite{CPYHKU:FAST-Dynamic-Vision} achieve the rotational and translational compensation by IMU and depth camera respectively. 
We design a motion compensation approach to correct each raw event position, which uses the angular velocity from the IMU sensor and the linear velocity from our ESVIO back-end to achieve rotational and translational compensation respectively.
Since our motion compensation utilizes the estimated velocity from the back-end, it works after the successful initialization.
%%%%%%%%%%%%%%%%%%***********************************************************************************************************************************************************%%%%%%%%%%%%%%%%%%
\begin{figure}[htb]  %%(h 此处（here） t 页顶（top）b 页底（bottom） p 独立一页（page）)
        \vspace{-2.0em}%调整表格与正文的距离
        \centering
        \captionsetup{justification=justified}%图题对齐
        \includegraphics[width=0.9\columnwidth]{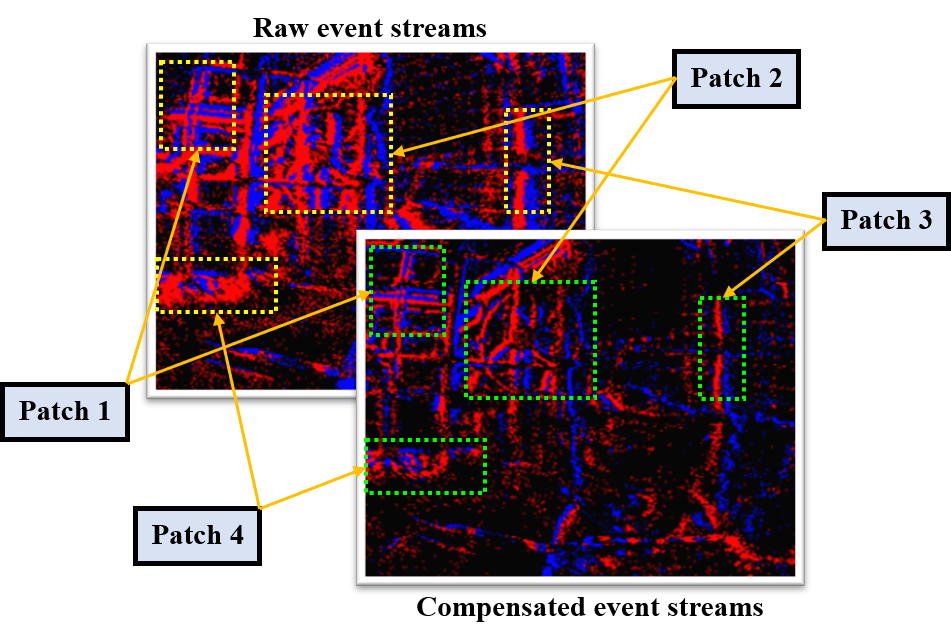}
        \caption{The event streams without and with motion compensation.}  %设置图片的一个编号以及为图片添加标题
        \label{motion_compensation}
        \vspace{-1.0em}%调整表格与正文的距离
\end{figure}%
%%%%%%%%%%%%%%%%%%***********************************************************************************************************************************************************%%%%%%%%%%%%%%%%%%

Given the $k$th event as $e_{k}=\{l_{k},t_{k},p_{k}\}$, where $l_{k}=\{x_{k},y_{k}\}$ represents the pixel location of the event $e_{k}$. $t_{k}$ is the timestamp and $p_{k}$ represents its polarity.
The event $e_{k}$ is warped from $t_{k}$ to $t_{ref}$, and the compensated location $^{ref}l_{k}$ define as
\vspace{-0.3cm}
\begin{equation}
\begin{split}
\label{compensated_location}
^{ref}l_{k} = \mathcal{M} [x_{k},y_{k},t_{k},\Theta]
\end{split}
\end{equation}
where $\mathcal{M}$ is the motion compensation function. $\Theta$ represents compensation parameters.
Since the short time interval $\Delta t$ between $t_{ref}$ and $t_{k}$, we assume that the motion during this period is uniform motion, thereby the ego-motion estimation of each event can be formulated as follow:
$$
\begin{aligned}
^{ref}\textbf{R}_{k} &= \textbf{\text{exp}}\big( (\boldsymbol{\tilde{\omega}}_{k}-\textbf{b}_{g}(t_{k})-\textbf{n}_{g}(t_{k}))\Delta t \big)\\
\end{aligned}
% \vspace{-0.3cm}
$$
\begin{equation}
\begin{split}
\label{motion_compensation_eq}
% ^{ref}\textbf{R}_{k} &= \textbf{\text{exp}}\big( (\boldsymbol{\tilde{\omega}}_{k}-\textbf{b}_{g}(t_{k})-\textbf{n}_{g}(t_{k}))\Delta t \big) \\
^{ref}\textbf{L}_{k} & =\hspace{0.035cm} ^{ref}\textbf{R}_{k} \textbf{L}_{k} + \textbf{v}_{ref}\Delta t \quad\quad\quad\quad\quad\hspace{0.05cm}\\
\end{split}
\end{equation}
where $\textbf{exp}$ denotes the exponential map $se(3)$$\to$$SE(3)$. $^{ref}\textbf{R}_{k}$ is the rotation matrix converted from the Euler angle $\boldsymbol{\omega_{k}}\Delta t$, $\boldsymbol{\omega}_{k}= \boldsymbol{\tilde{\omega}}_{k}-\textbf{b}_{g}(t_{k})-\textbf{n}_{g}(t_{k})$. $\boldsymbol{\tilde{\omega}}_{k}$ is the measurement of gyroscope, while $\textbf{b}_{g}(t_{k})$ and $\textbf{n}_{g}(t_{k})$ are bias and noise variable of gyroscope respectively. $\textbf{L}_{k}=\{x_{k},y_{k},1\}$ is homogeneous matrix that extended from $l_{k}$. $\textbf{v}_{ref}$ represents the velocity of our ESVIO back-end at $t_{ref}$ timestamp. Finally, we convert $^{ref}\boldsymbol{L}_{k}$ to homogeneous matrix and obtain the compensated location $^{ref}l_{k}$.
% Meanwhile, motion compensation only performs on event streams when the acceleration of our back-end exceeds the preset threshold.
Fig. \ref{motion_compensation} compares raw event streams and motion-compensated event streams, where raw event streams produce a certain extent of distortion while the compensated event streams show clear contours of scenes.

%%%%%%%%%%%%%%%%%%%%%%%%%%%%%%%%%%%%%%%%%%%%%%%%%%%%%%%%%%%%%%%%%%%%%%%%%%%%%%%%
\subsection{Event-based Spatial and Temporal Data Associations}
\label{Event-based Spatial and Temporal Data Associations}

After the motion compensation, the stereo event streams are fed to generate two (positive and negative) surface-of-active-event (SAE) which store the event pixels and timestamps.
For the new arrival event streams, the existing event-corner features are firstly temporally tracked by the LK optical approach \cite{CPYHKU:LK_optical_flow} and then spatially matched in left and right event streams.
The event-corner features that are not successfully tracked and matched in the current timestamp would be discarded immediately.
While new event-corner features are extracted on the motion-compensated event streams to maintain a minimum number (100-200) of features in each timestamp.
Modified from the publicly available implementation of the Arc* algorithm \cite{CPYHKU:ARC*} for event-based corner detection, we extract the event corners on the individual event by leveraging the SAE.
We only select those events whose timestamps are within a short interval from that of the current time surface, thereby retaining event-corner features at the dense event streams.
% To reduce the influence of noisy events, we further apply a mask to filter effective event-corner features.
% The mask is a time surface (TS) with polarity \cite{CPYHKU:GuanEVIO}, which is also used to distribute adjacent event-corner features uniformly by setting a minimum distance $d_{min}$ value.
To reduce the influence of noisy events, we further apply time surface (TS) with polarity as a mask to filter effective event-corner features, and the TS is converted from the SAE in real time.
Meanwhile, TS is also used to distribute adjacent event-corner features uniformly by setting a minimum distance $d_{min}$ value.

%%%%%%%%%%%%%%***********************************************************************************************************************************************************%%%%%%%%%%%%%%%%%%
\begin{figure}[htb]  %%(h 此处（here） t 页顶（top）b 页底（bottom） p 独立一页（page）)
        \vspace{-1.0em}%调整表格与正文的距离
        \centering
        \captionsetup{justification=justified}%图题对齐
        \includegraphics[width=1.0\columnwidth]{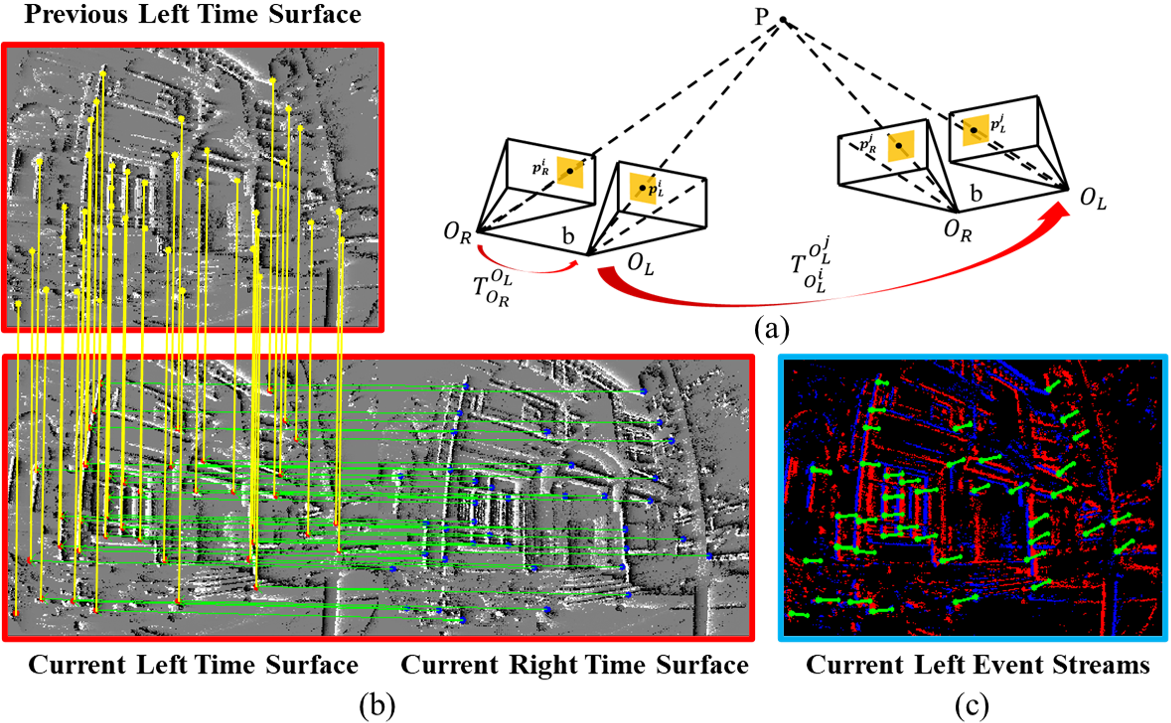}
        \caption{Stereo event-corner features:
    (a) Geometry principle;
    (b) Temporally and spatially associating the event-corner features on the time surface;
    (c) Event-corner features tracking on the event streams. }  %设置图片的一个编号以及为图片添加标题
        \label{tracking}
        \vspace{-1.0em}%调整表格与正文的距离
\end{figure}%
%%%%%%%%%%%%%%%%%%***********************************************************************************************************************************************************%%%%%%%%%%%%%%%%%%
%%%%%%%%%%%%%%%%%%%%这一段讲temporal和instantaneous匹配过程%%%%%%%%%%%%%%%%%%%5
% To obtain the 3D coordinate of event-corner features and perform nonlinear optimization, we adopt both temporal and spatial event-based associations. 
The geometry principle of temporal and spatial event-based associations is depicted in Fig. \ref{tracking}(a).
To ensure that the matched features between the left and right event streams lie along the epipolar line, we instantaneously match stereo-rectified time surfaces for the spatial association.
Stereo event-corner features, $\mathcal{F}_{l}^{i}$ and $\mathcal{F}_{r}^{i}$, are instantaneously matched by forward and inverse LK optical flow between the left and right time surface at the current timestamp $i$. 
Our ESVIO executes spatial and temporal association at each frame.
Meanwhile, the temporal association also uses forward and inverse optical flow to track event-corner features of left event streams, $\mathcal{F}_{l}^{i}$ and $\mathcal{F}_{l}^{j}$, at consecutive two timestamps $i$ and $j$. 
In Fig. \ref{tracking}(b), red and blue dots represent event-corner features extracted on the left and right time surface at timestamp $i$ respectively, while yellow dots denote the features on the left time surface at the previous timestamp $j$.
Note that the left bottom and right bottom pictures form the spatial event-based association, while the left bottom and left top pictures form the temporal event-based association. 
Green lines connect matched event-corner features between the left and right time surfaces at the same timestamp, while yellow lines connect tracked features with two consecutive left time surfaces at $i$ and $j$.
Fig. \ref{tracking}(c) and the bottom of Fig. \ref{firstpic} show our proposed method can achieve correct temporal and spatial association in consecutive stereo event-corner features, even when the image-based tracking is failed caused by the motion blur.

% Moreover, the Random Sample Consensus(RANSAC) is used to eliminate the outliers. 

%%%%%%%%%%%%%%%%%%%%%这里是讲如何三角化得到特征点3D位置的，上面是讲如何实现特征点匹配的%%%%%%%%%%%%%%%
Finally, we recover the inverse depth of event-corner features by RANSAC outlier rejection and triangulation.
As depicted in Fig. \ref{tracking}(a), the matched corner features of the 3D point $P$ on the imaging plane of left and right event cameras at timestamp $i$ are $p_{L}^{i}$ and $p_{R}^{i}$ respectively. The inverse depth of $P$ can be formulated through epipolar geometry and triangulation between $p_{L}^{i}$ and $p_{R}^{i}$.
Similarly, $p_{L}^{i}$ and $p_{L}^{j}$ can also be used to recover the inverse depth.
Based on these associated event-corner features, corresponding residual constraints can be constructed to conduct graph-based optimization.
% Meanwhile, the event-corner feature of left event camera $p_{L}^{i}$ and $p_{L}^{j}$ at timestamps $i$ and $j$ can also be triangulated to recover the inverse depth.
% $O_{L},O_{R}$ represents the aperture of the left and right event cameras, respectively.
% $b$ is the baseline of the stereo event camera. 
% $T_{O_{R}}^{O_{L}}$ is the homogeneous transformation from the right event camera to the left event camera, while $T_{O_{L}^{i-th}}^{O_{L}^{j-th}}$ is the movement of left event camera from timestamp $i$th to $j$th.

%%%%%%%%%%%%%%%%%%%%%%%%%%%%%%%%%%%%%%%%%%%%%%%%%%%%%%%%%%%%%%%%%%%%%%%%%%%%%%%%
\subsection{The Construction of Event-based Residual Constraint for the Graph-based Optimization} 
\label{The Construction of Event-based Residual Constraint for the Graph-based Optimization} 
The full state vector in the sliding window is defined as
$$
\begin{aligned}
\boldsymbol{\chi} &= \left[\textbf{x}_{b_{0}},...,\textbf{x}_{b_{n}}, \textbf{x}^{b}_{e}, \textbf{x}^{b}_{c}, \boldsymbol{\Lambda}_{es},\boldsymbol{\Lambda}_{et},
\boldsymbol{\Lambda}_{c}\right]\hspace{0.15cm}\\
\end{aligned}
$$
\begin{equation}
\begin{split}
\label{state}
% \boldsymbol{\chi} &= \left[\textbf{x}_{b_{0}},...,\textbf{x}_{b_{n}}, \textbf{x}^{b}_{e}, \textbf{x}^{b}_{c}, \boldsymbol{\Lambda}_{et},
% \boldsymbol{\Lambda}_{em},
% \boldsymbol{\Lambda}_{c}\right]\\
\textbf{x}_{b_{k}} &= \left[ \textbf{p}_{b_{k}}^{w}, \textbf{q}_{b_{k}}^{w},\textbf{v}_{b_{k}}^{w},\textbf{b}_{a_{k}},\textbf{b}_{g_{k}} \right] \quad k\in [0,n]
% \Lambda_{et} &= [\lambda_{et_{0}},\lambda_{et_{1}},...,\lambda_{et_{l}}]\\
% \Lambda_{em} &= [\lambda_{em_{0}},\lambda_{em_{1}},...,\lambda_{em_{l}}]\\
% \Lambda_{c} &= [\lambda_{c_{0}},\lambda_{c_{1}},...,\lambda_{c_{l}}]
% x^{b}_{e}&=[ p_{e}^{b}, q_{e}^{b}] \quad x^{b}_{c}=[ p_{c}^{b}, q_{c}^{b}]\\
\end{split}
\end{equation}
where $\textbf{x}_{b_{k}}$ is the state of IMU at timestamp $k$ in the world frame, which consists of the position $\textbf{p}_{b_{k}}^{w}$, the orientation quaternion $\textbf{q}_{b_{k}}^{w}$, the velocity $\textbf{v}_{b_{k}}^{w}$, the accelerometer bias $\textbf{b}_{a_{k}}$ and the gyroscope bias $\textbf{b}_{g_{k}}$. $\textbf{x}^{b}_{e}$ and $\textbf{x}^{b}_{c}$ are the extrinsic transformation from event cameras and standard cameras to IMU respectively. $\boldsymbol{\Lambda}_{es} = [\lambda_{es_{0}},\lambda_{es_{1}},...,\lambda_{es_{l}}]$,
$\boldsymbol{\Lambda}_{et} = [\lambda_{et_{0}},\lambda_{et_{1}},...,\lambda_{et_{l}}]$,  $\lambda_{es_{l}}, \lambda_{et_{l}}$ represent the inverse depth of the event-corner features $es_{l}$, $et_{l}$ respectively.
$\boldsymbol{\Lambda}_{c} = [\lambda_{c_{0}},\lambda_{c_{1}},...,\lambda_{c_{l}}]$, $\lambda_{c_{l}}$ is the inverse depth of the image-based features $c_{l}$. $n$ is the total number of keyframes, and $l$ is the total number of features in the sliding window.

Combining event, image, and IMU residual terms, the event-visual-inertial odometry can be formulated as the joint nonlinear optimization problem as follow
\begin{equation}
\begin{split}
\label{Joint_nonlinear_optimization}
\min\limits_{\boldsymbol{\chi}} &\Bigg(
% \left \|r_{m}\right \|^{2}_{W_{m}} + 
% \left \|r_{r}\right \| ^{2}_{W_{r}}+
\sum_{k \in b}\left \|\textbf{r}_{b}(\hat{\textbf{z}}_{b_{k+1}}^{b_{k}}, \boldsymbol{\chi})\right \|^{2}_{\Omega_{b}}
+\sum_{(l,k) \in es}\left \|\textbf{r}_{es}(\hat{\textbf{z}}_{es_{k}}^{l}, \boldsymbol{\chi})\right \|^{2}_{\Omega_{es}} \\
&+\sum_{(l,k) \in et}\left \|\textbf{r}_{et}(\hat{\textbf{z}}_{et_{k}}^{l}, \boldsymbol{\chi})\right \|^{2}_{\Omega_{et}}
+ \sum_{(l,k) \in c}\left \|\textbf{r}_{c}(\hat{\textbf{z}}_{c_{k}}^{l}, \boldsymbol{\chi})\right \|^{2}_{\Omega_{c}} \Bigg)
\end{split}
\end{equation}
% where $r_{m}$ is the marginalization residual with weight $W_{m}$. The relocalization residual $r_{r}$ with weight $W_{r}$. 
where $\textbf{r}_{b}(\hat{\textbf{z}}_{b_{k+1}}^{b_{k}}, \boldsymbol{\chi})$ is the residual for IMU measurement with information matrix $\Omega_{b}$. $\textbf{r}_{es}(\hat{\textbf{z}}_{es_{k}}^{l}, \boldsymbol{\chi})$ and $\textbf{r}_{et}(\hat{\textbf{z}}_{et_{k}}^{l}, \boldsymbol{\chi})$ are the residuals for event-based spatial and temporal association measurement, with corresponding information matrix $\Omega_{es}$ and $\Omega_{et}$ respectively. $\textbf{r}_{c}(\hat{\textbf{z}}_{c_{k}}^{l}, \boldsymbol{\chi})$ represents the residuals for standard cameras measurement with information matrix $\Omega_{c}$. 
The definition of event-based residuals will be presented below, while other residual terms can be found in \cite{CPYHKU:PL-EVIO}\cite{CPYHKU:VINS-Fusion}.

% \vspace{-1.0em}
For the stereo event cameras, we construct the spatial association factor $\textbf{r}_{es}(\hat{\textbf{z}}_{es_{k}}^{l}, \boldsymbol{\chi})$ in the consecutive event streams and the temporal association factor $\textbf{r}_{et}(\hat{\textbf{z}}_{et_{k}}^{l}, \boldsymbol{\chi})$ between the left and right event streams.
Consider the $l$th feature that is observed in the $i$th right event stream, the residual for the event-corner feature observation in the $i$th left event stream is defined as:
\begin{equation}
% \vspace{-0.6cm} 
\begin{split}
\setlength{\arraycolsep}{1.2pt}
        \textbf{r}_{es}=\left[
                \begin{array}{c}
                u^{l}_{les_{i}}\\
                v^{l}_{les_{i}}\\
                \end{array} 
                \right]-\pi_{e}\bigg(\textbf{T}^{le}_{re} \pi^{-1}_{e}(\frac{1}{\lambda_{es}},\left[
                        \begin{array}{c}
                        u^{l}_{res_{i}}\\
                        v^{l}_{res_{i}}\\
                        \end{array} 
                        \right])\bigg)
 \label{re-projection-event-instantaneous-matching}
\end{split}
\end{equation}
where $[u^{l}_{les_{i}}, v^{l}_{les_{i}}]$ is the observation of the $l$th event-corner feature in the $i$th left event stream. $[u^{l}_{res_{i}}, v^{l}_{res_{i}}]$ is the same event-corner feature in the $i$th right event stream.
$\pi_{e}$ and $\pi^{-1}_{e}$ are the projection and back-projection functions of the event camera respectively.
$\textbf{T}_{re}^{le}$ represents the extrinsic transformation from the right to left event camera. 

Consider the $l$th feature that is first observed in the $i$th left event stream, the residual for the event-corner feature observation in the $k$th left event stream is defined as:
\begin{equation}
\vspace{-0.2cm} 
\begin{split}
\setlength{\arraycolsep}{1.2pt}
        \textbf{r}_{et}=\left[
                \begin{array}{c}
                u^{l}_{et_{k}}\\
                v^{l}_{et_{k}}\\
                \end{array} 
                \right]-\pi_{e}\bigg((\textbf{T}^{b}_{le})^{-1} \textbf{T}^{b_{k}}_{w} \textbf{T}^{w}_{b_{i}} \textbf{T}^{b}_{le} \pi^{-1}_{e}(\frac{1}{\lambda_{et}},\left[
                        \begin{array}{c}
                        u^{l}_{et_{i}}\\
                        v^{l}_{et_{i}}\\
                        \end{array} 
                        \right])\bigg)
 \label{re-projection-event-temporal-tracking}
\end{split}
\end{equation}
where $[u^{l}_{e_{k}}, v^{l}_{e_{k}}]$ is the observation of the $l$th event-corner feature in the $k$th event stream. $[u^{l}_{e_{i}}, v^{l}_{e_{i}}]$ is the same event-corner feature in the $i$th event stream. 
% $T$ is the 4x4 homogeneous transformation. 
$\textbf{T}_{le}^{b}$ represents the extrinsic transformation from the left event camera to the body coordinate.
$\textbf{T}^{w}_{b_{i}}$ indicates the pose of the body center related to the world frame at timestamp $i$, $\textbf{T}^{b_{k}}_{w}$ is the transpose of the pose of the body coordinate in the world frame at the $k$th keyframe.
%%%%%%%%%%%%%%%%%%%%%%%%%%%%%%%%%%%%%%%%%%%%%%%%%%%%%%%%%%%%%%%%%%%%%%%%%%%%%%%%%

%%%%%%%%%%%%%%%%%%%%%%%%%%%%%%%%%%%%%%%%%%%%%%%%%%%%%%%%%%%%%%%%%%%%%%%%%%%%%%%%
\section{Evaluation}
\label{Evaluation}

We perform both dataset and real-world experiments to evaluate our proposed methods.
We first evaluate our proposed ESIO and ESVIO in the self-collected dataset which is acquired by two DAVIS346 ($346 \times 260$, event-sensor, image-sensor, IMU sensor) and VICON.
It contains extremely fast 6-Dof motion and scenes with HDR. 
In subsection \ref{Evaluation of Our ESVIO on Public Datasets}, we compare our methods with other event-based and image-based methods on two publicly available datasets: MVSEC \cite{CPYHKU:MVSEC} and VECtor \cite{CPYHKU:VECtor}.
We perform quantitative analysis to evaluate the accuracy of our system.
The accuracy is measured with mean position error (MPE, \%) and mean rotation error (MRE, $^\circ$/m) aligning the estimated trajectory with ground truth using 6-DOF transformation (in SE3), which is calculated by the tool \cite{CPYHKU:evo_package}.
Finally, in subsection \ref{Indoor Quadrotor Flighting Evaluation} we evaluate our ESVIO in the onboard quadrotor flighting. 
While subsections \ref{DSEC} and \ref{outdoor large-scale} perform the evaluation of the autonomous-driving dataset and outdoor large-scale environment, respectively.
All experiments run in real-time on an Intel NUC computer equipped with Intel i7-1260P, 32GB RAM, and Ubuntu 20.04 operation system.

%%%%%%%%%%%%%%%%%%***********************************************************************************************************************************************************%%%%%%%%%%%%%%%%%%
\begin{figure}[htb]  %%(h 此处（here） t 页顶（top）b 页底（bottom） p 独立一页（page）)
        \setlength{\abovecaptionskip}{-0.5em}%调整图片标题与图距离
        \vspace{-2.0em}%调整表格与正文的距离
        \centering
        \captionsetup{justification=justified}%图题对齐
        \includegraphics[width=0.85\columnwidth]{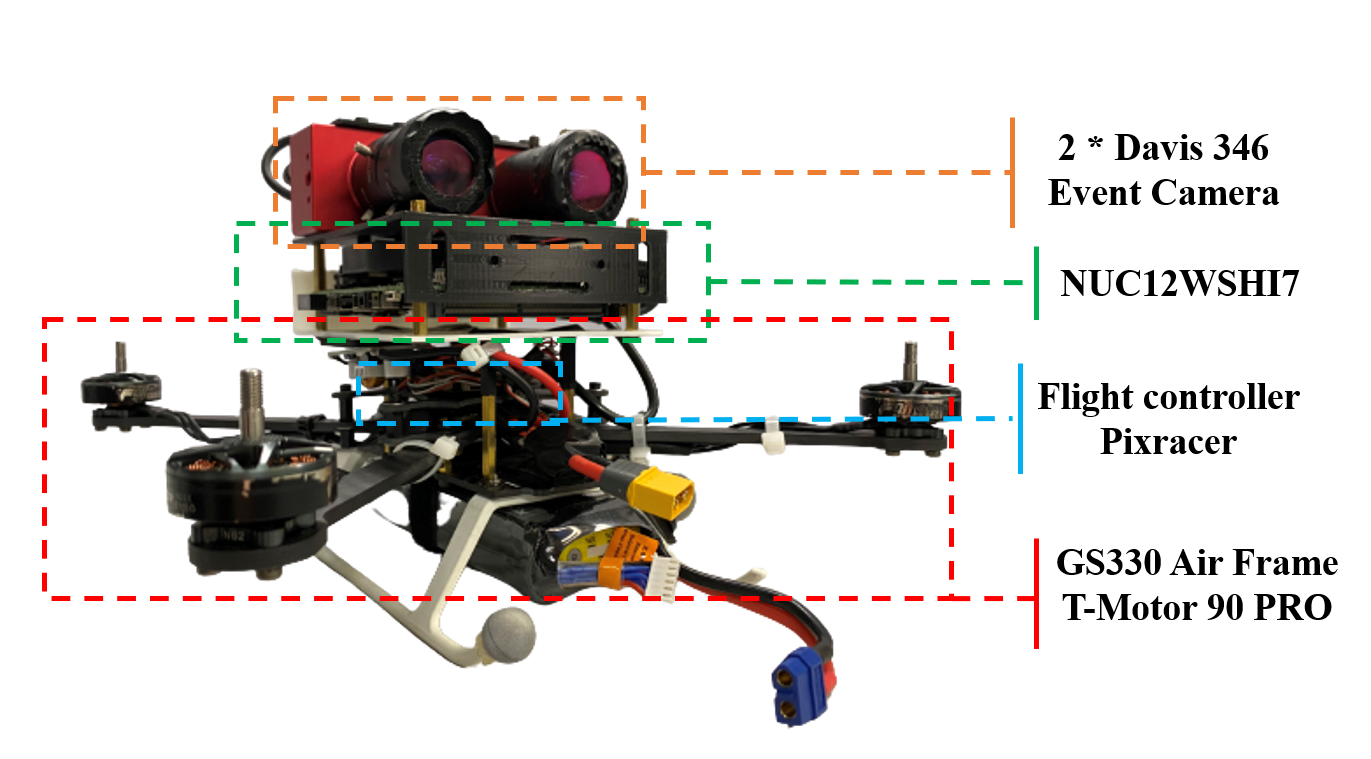}
        \caption{Our self-designed quadrotor platform.}  %设置图片的一个编号以及为图片添加标题
        \label{quadrotor_flight}
        \vspace{-2.0em}%调整表格与正文的距离
\end{figure}%
%%%%%%%%%%%%%%%%%%***********************************************************************************************************************************************************%%%%%%%%%%%%%%%%%%

%%%%%%%%%%%%%%%%%%%%%%%%%%%%%%%%%%%%%%%%%%%%%%%%%%%%%%%%%%%%%%%%%%%%%%%%%%%%%%%%
\subsection {Evaluation of Our ESVIO in Challenging Situations}
\label{Evaluation of Our ESVIO in Challenge Situation}

%%%%%%%%%%%%%%%%%%%%%%%%%%%%%%%%%%%%%%%%%%%%%%%%%%%%%%%%%%%%%%%%%%%%%%%%%%%%%%%%
\begin{table*}[htbp]
% \begin{adjustbox}{minipage=\linewidth,bgcolor=yellow}
        % \vspace{-1.0em}%调整表格与正文的距离
        % \setlength{\abovecaptionskip}{-0.06cm}%调整图片标题与图距离
        \renewcommand\arraystretch{1.2}
        % \linespread{1.6}
        \begin{center}
        \caption{The Accuracy Comparison of Our ESVIO with other Image-based or Event-based Methods on HKU Dataset}
        \label{vicon_Comparison_with_ESVIO}
        \resizebox{\columnwidth*2}{!}
        { 
        \begin{threeparttable}
        \begin{tabular}{c|cccccccc} 
%tabular环境是LaTeX默认创建表格的环境。你需要对这个环境手动指定一个参数。{c c c}参数告诉LaTeX，表格将会有三列，每一列都是居中对齐（c: center）
        \hline  %这个参数会在表格中插入水平的分割线。你可以多次使用这个命令。
        \multicolumn{1}{c|}{\multirow{2}*{Sequence}} 
    & \makecell{ORB-SLAM3 \cite{CPYHKU:ORB-SLAM3} \\Stereo VIO}
    & \makecell{VINS-Fusion \cite{CPYHKU:VINS-Fusion} \\Stereo VIO}
    % & \makecell{EVO \cite{CPYHKU:EVO} \\Mono EO} 
    % & \makecell{ESVO \cite{CPYHKU:ESVO} \\Stereo EO} 
    &\makecell{USLAM \cite{CPYHKU:ETH-EIO} \\Mono EIO} 
    &\makecell{USLAM \cite{CPYHKU:Ultimate-SLAM} \\Mono EVIO}
    &\makecell{PL-EVIO \cite{CPYHKU:PL-EVIO} \\Mono EVIO}   
    & \makecell{\textbf{Our ESIO} \\\textbf{Stereo EIO}}
    & \makecell{\textbf{Our ESIO+}  \\\textbf{Stereo EIO}}
    &  \makecell{\textbf{Our ESVIO} \\\textbf{Stereo EVIO}}\\
    \cline{2-9}
    & MPE / MRE & MPE / MRE & MPE / MRE & MPE / MRE & MPE / MRE & MPE / MRE & MPE / MRE & MPE / MRE
    \\   
    % &  Our ESVIO \cite{GWPHKU:MyEVIO} 
    % &  PL-EVIO \\  %&符号分割了单元格之间的内容。\\代表着一行的结束。
\hline
hku\_agg\_translation            & 0.15 / 0.075& 0.11 / 0.019& 16.22 / 0.45& 0.59 / 0.020& \textbf{0.07} / 0.091& 0.59 / 0.16& 0.55 / 0.16&0.10 / \textbf{0.016}\\
hku\_agg\_rotation             & 0.35 / 0.11& 1.34 / 0.024& \textit{failed}& 3.14 / 0.026& 0.23 / 0.12& 1.33 / 0.048& 0.78 / 0.045& \textbf{0.17} / \textbf{0.015}\\
hku\_agg\_flip            & \textbf{0.36} / 0.39& 1.16 / 2.02& 11.15 / 2.11& 6.86 / 2.04& 0.39 / 2.23& 3.79 / 0.23& 3.17 / 0.23& \textbf{0.36} / \textbf{0.12}\\
hku\_agg\_walk            & \textit{failed} &\textit{failed}& \textit{failed}& 2.00 / 0.16& 0.42 / 0.14& 1.49 / 0.23& 1.30 / 0.23& \textbf{0.31} / \textbf{0.026}\\
hku\_hdr\_circle    & 0.17 / 0.12& 5.03 / 0.60& 0.92 / 0.58& 1.32 / 0.54& \textbf{0.14} / 0.62& 1.38 / 0.10& 0.46 / 0.099& 0.16 / \textbf{0.035}\\
hku\_hdr\_slow    & 0.16 / 0.058& 0.13 / \textbf{0.026}& \textit{failed}& 2.80 / 0.099& 0.13 / 0.068& 0.29 / 0.38& 0.31 / 0.39& \textbf{0.11} / 0.028\\
hku\_hdr\_tran\_rota    & 0.30 / 0.042& 0.11 / 0.021& \textit{failed}& 2.64 / 0.13& 0.10 / 0.064& 0.84 / 0.30& 0.91 / 0.31& \textbf{0.10} / \textbf{0.018}\\
hku\_hdr\_agg    & 0.29 / 0.085& 1.21 / 0.27& \textit{failed}& 2.47 / 0.27& 0.14 / 0.30& 2.33 / 0.16& 1.41 / 0.14& \textbf{0.10} / \textbf{0.021}\\
hku\_dark\_normal           & \textit{failed}& 0.86 / 0.028& \textit{failed}& 2.17 / 0.031& 1.35 / 0.081& \textbf{0.30} / 0.12& 0.35 / 0.12& 0.42 / \textbf{0.015}\\
\hline 
Average                & 0.16 / 0.12& 0.76 / 0.38& 5.06 / 1.05& 1.69 / 0.39& 0.26 / 0.41& 0.89 / 0.19& 0.66 / 0.19& \textbf{0.14} / \textbf{0.033} \\
\hline        
        \end{tabular}
        \begin{tablenotes} 
        \item \textit{*EIO means purely event-based VIO, EVIO means event-based VIO with image-aided} 
        \end{tablenotes} 
        \end{threeparttable} 
        }
        \end{center}
% \end{adjustbox}
        \vspace{-2.5em}
\end{table*}
%%%%%%%%%%%%%%%%%%%%%%%%%%%%%%%%%%%%%%%%%%%%%%%%%%%%%%%%%%%%%%%%%%%%%%%%%%%%%%%%%%%%%%%%%%%%%%%%%%%%%%%%%%%%

%%%%%%%%%%%%%%%%%%%%%%%%%%%%%%%%%%%%%%%%%%%%%%%%%%%%%%%%%%%%%%%%%%%%%%%%%%%%%%%%
\subsubsection{\bf Experiment Data Description}

% In this section, we describe the datasets that we provided. 
The dataset contains stereo event data at 60Hz and stereo image frames at 30Hz with resolution in $346 \times 260$, as well as IMU data at 1000Hz.
Timestamps between all sensors are synchronized in hardware.
We also provide ground truth poses from a motion capture system VICON at 50Hz for each sequence, which can be used for accuracy comparison.
The dataset consists of handheld sequences including rapid motion and HDR scenarios.
% To alleviate disturbance from the motion capture system’s infrared light on the event camera, we add an infrared filter on the lens surface of the DAVIS346 camera. 
% Note that the introduction of the infrared filter might cause the degradation of perception for both the event and image camera during the evaluation, but this can further increase the difficulty of our dataset and show the robustness of our proposed method.
The full setup including the attached infrared filter can be seen in Fig. \ref{quadrotor_flight}.
% (the details of the quadrotor platform would be introduced in Section \ref{Indoor Quadrotor Flighting Evaluation}).
These two DAVIS346 are rigidly attached with a baseline of 6.0 cm and USB 3.0 interfaces are used to transmit sensor measurements to the NUC.
However, since the limitation of our hardware and cost, we use DAVIS346-COLOR and DAVIS346-MONO for the data collection.
Although this might introduce some artificial inconsistency, we think it is acceptable for the method evaluation.
The DAVIS comprises an image camera and event camera on the same pixel array, thus calibration can be done using standard image-based methods, such as Kalibr\footnote{\url{https://github.com/ethz-asl/kalibr}}, on the image frames and then are applied to the event camera.
For the benefit of the research community, we also release the dataset and the configuration files on our project website.
% While the detail description of each sequence is also available in the website.

%%%%%%%%%%%%%%%%%%%%%%%%%%%%%%%%%%%%%%%%%%%%%%%%%%%%%%%%%%%%%%%%%%%%%%%%%%%%%%%%
\subsubsection{\bf Methods Evaluation}
Table \ref{vicon_Comparison_with_ESVIO} compares the performance of our ESIO and ESVIO with the other state-of-the-art event-based or image-based systems.
Our ESIO has good performance, especially for the sequence \textit{hku\_agg\_walk} and \textit{hku\_dark\_normal}, our ESIO still can produce reliable and accurate pose estimation even when the state-of-the-art image-based VIO method, ORB-SLAM3, fails.
% Due to motion blur and poor light conditions, ORB-SLAM3 fails to extract and track valid ORB features in \textit{hku\_agg\_walk} and \textit{hku\_dark\_normal} sequences, resulting in system failure (more details can be seen in the supplementary material).
Due to motion blur, both ORB-SLAM3 and VINS-Fusion fail to extract reliable features in \textit{hku\_agg\_walk} sequence, resulting in system failure. In the \textit{hku\_dark\_normal} sequence, ORB-SLAM3 cannot extract any feature due to poor light conditions (more qualitative details is in the supplementary material).
As for the MRE evaluation criterion, our ESVIO shows significant improvement compared to other advanced algorithms, e.g. the average MRE of ESVIO is 0.033$^\circ$/m while the value of ORB-SLAM3 is 0.12$^\circ$/m.
Even if it does not perform too much improvement in MPE compared with our previous work PL-EVIO \cite{CPYHKU:PL-EVIO} in most sequences, ESIO and ESVIO are still a breakthrough for event-based stereo VIO.
For the motion compensation version (ESIO+), it shows effective improvement in most of the data sequences compared to ESIO, e.g. the average MPE of ESIO+ is 0.66\% compared to 0.89\% of ESIO, which is opposite to the conclusion from Ref.\cite{CPYHKU:PL-EVIO}.
This might be thanks to the well-designed motion-compensated methods that process the reliable and optimized compensated measurements from the ESVIO back-end.
% Furthermore, we also evaluate our ESVIO which has significant performance improvement compared with the ESIO. 

Note that we also evaluate EVO\cite{CPYHKU:EVO} and ESVO\cite{CPYHKU:ESVO} in our self-collected datasets, but they failed in all sequences.
This might be caused by three factors: 
Firstly, both EVO and ESVO have strict initialization requirements. For example, EVO requires running in a uniform scene for a few seconds to boost the system.
Secondly, they are sensitive to parameter tuning, even in their open-source project, they use different parameters for different sequences in the same scenarios. We might fail to correctly tune parameters for their successful running.
Finally, our dataset is so challenging that only reliable methods can perform well.

% \vspace{-0.5em}
%%%%%%%%%%%%%%%%%%%%%%%%%%%%%%%%%%%%%%%%%%%%%%%%%%%%%%%%%%%%%%%%%%%%%%%%%%%%%%%%
\subsection{Evaluation of Our ESVIO on Public Datasets}
\label{Evaluation of Our ESVIO on Public Datasets}

In this section, we evaluate our ESVIO on publicly available datasets. 
The VECtor \cite{CPYHKU:VECtor} dataset consists of a hardware-synchronized sensor suite that includes stereo event cameras, stereo standard cameras, an RGB-D sensor, a LiDAR, and an IMU.
It covers the full spectrum of 6 DoF motion dynamics, environment complexities, and illumination conditions for both small and large-scale scenarios.
To the best of our knowledge, we provide the first results on this new event-based dataset.
For the MVSEC \cite{CPYHKU:MVSEC}, we select the sequence captured in the indoor flying room.
We use the stereo event camera ($640 \times 480$) and the regular stereo camera ($1224 \times 1024$) from the VECtor, and the DAVIS346 ($346 \times 260$ for both event and image) from the MVSEC, for evaluation, respectively.

% %%%%%%%%%%%%%%%%%%%%%%%%%%%%%%%%%%%%%%%%%%%%%%%%%%%%%%%%%%%%%%%%%%%%%%%%%%%%%%%%%%%%%%%%%%%%%%%%%%%%%%%%%%%%%%%%%%%%%%%%%%%%%%%%%%%%%%%%%%%%%%%%%%%%%%%%
\begin{table*}[htbp]
% \begin{adjustbox}{minipage=\linewidth,bgcolor=yellow}
        % \vspace{-1.0em}%调整表格与正文的距离
        % \setlength{\abovecaptionskip}{-0.06cm}%调整图片标题与图距离
        \renewcommand\arraystretch{1.2}
        %  \linespread{1.6}
        % \LARGE %此处写字体大小控制命令
        \begin{center}
        \caption{The Accuracy Comparison of Our ESVIO with other Image-based or Event-based Methods on Public Dataset}
        \label{public_dataset}
        \resizebox{\columnwidth*2}{!}
        { 
        \begin{threeparttable}
        \begin{tabular}{c|c|ccccccc} 
        \hline  
    \multicolumn{2}{c|}{\multirow{2}*{Sequence}}  
    & \makecell{ORB-SLAM3 \cite{CPYHKU:ORB-SLAM3} \\Stereo VIO} 
    & \makecell{VINS-Fusion \cite{CPYHKU:VINS-Fusion} \\Stereo VIO}
    & \makecell{EVO \cite{CPYHKU:EVO} \\Mono EO}
    & \makecell{ESVO \cite{CPYHKU:ESVO} \\Stereo EO} 
    & \makecell{Ultimate SLAM \cite{CPYHKU:Ultimate-SLAM}\\Mono EVIO } 
    & \makecell{PL-EVIO \cite{CPYHKU:PL-EVIO} \\Mono EVIO} 
    & \makecell{\textbf{Our ESVIO} \\\textbf{Stereo EVIO}}\\
    \cline{3-9}
    \multicolumn{2}{c|}{} & MPE / MRE & MPE / MRE & MPE / MRE & MPE / MRE & MPE / MRE & MPE / MRE & MPE / MRE \\
    %&符号分割了单元格之间的内容。\\代表着一行的结束。
\hline
\multirow{17}*{\makecell{VECtor\cite{CPYHKU:VECtor}}}
  & corner-slow  & 1.49 / 14.28& 1.61 / 14.06& 4.33 / 15.52 & 4.83 / 20.98& 4.83 / 14.42& 2.10 / 14.21& \textbf{1.49} / \textbf{14.03}\\
~ & robot-normal  & 0.73 / 1.18& \textbf{0.58} / 1.18& 3.25 / 2.00 &\textit{failed}& 1.18 / \textbf{1.11}& 0.68 / 1.25& 1.08 / 1.17\\
~ & robot-fast & 0.71 / 0.70& \textit{failed}& \textit{failed} &\textit{failed}& 1.65 / 0.56&  \textbf{0.17} / 0.74& 0.20 / \textbf{0.56}\\
~ & desk-normal  & \textbf{0.46} / 0.41& 0.47 / \textbf{0.36}& \textit{failed} &\textit{failed}& 2.24 / 0.56 & 3.66 / 0.45& 0.61 / 0.38\\
~ & desk-fast  & 0.31 / 0.41& 0.32 / 0.33& \textit{failed} &\textit{failed} &  1.08 / 0.38 & 0.14 / 0.48& \textbf{0.13} / \textbf{0.32}\\
~ & sofa-normal  & \textbf{0.15} / 0.41& 0.13 / 0.40& \textit{failed} &1.77 / 0.60& 5.74 / \textbf{0.39}&  0.19 / 0.46& 0.16 / 0.40\\
~ & sofa-fast  & 0.21 / 0.43& 0.57 / \textbf{0.34}& \textit{failed} &\textit{failed}& 2.54 / 0.36 & \textbf{0.17} / 0.47& 0.17 / 0.35\\
~ & mountain-normal  & \textbf{0.35} / 1.00& 4.05 / 1.05& \textit{failed} & \textit{failed}& 3.64 / 1.06& 4.32 / \textbf{0.76}& 0.59 / 0.77\\
~ & mountain-fast  & 2.11 / 0.64& \textit{failed} & \textit{failed} &\textit{failed}& 4.13 / 0.62& \textbf{0.13} / 0.56& 0.16 / \textbf{0.45}\\
~ & hdr-normal  &0.64 / 1.20& 1.27 / 1.10& \textit{failed} &\textit{failed} & 5.69 / 1.65&  4.02 / 1.52& \textbf{0.57} / \textbf{1.06}\\
~ & hdr-fast  & 0.22 / 0.45& 0.30 / 0.34& \textit{failed} &\textit{failed} & 2.61 / 0.34& \textbf{0.20} / 0.50& 0.21 / \textbf{0.33}\\
~ & corridors-dolly &\textbf{1.03} / 1.37& 1.88 / 1.37& \textit{failed} &\textit{failed}  & \textit{failed} & 1.58 / 1.37& 1.13 / \textbf{1.33} \\
~ & corridors-walk  &1.32 / 1.31& 0.50 / 1.31& \textit{failed} &\textit{failed}  & \textit{failed} & 0.92 / \textbf{1.31}& \textbf{0.43} / 1.32\\
~ & school-dolly     &0.73 / 1.02&1.42 / 1.06 & \textit{failed} &10.87 / 1.08  & \textit{failed}  & 2.47 / 0.97& \textbf{0.42} / \textbf{0.73}\\
~ & school-scooter   &0.70 / \textbf{0.49}& \textbf{0.52} / 0.61& \textit{failed} &9.21 / 0.63 & 6.40 / 0.61& 1.30 / 0.54& 0.59 / 0.56\\
~ & units-dolly    &7.64 / 0.41& 4.39 / 0.42& \textit{failed} &\textit{failed}  & \textit{failed} & 5.84 / 0.44& \textbf{3.43} / \textbf{0.022}\\
~ & units-scooter  &6.22 / \textbf{0.22}& 4.92 / 0.24& \textit{failed} &\textit{failed}  & \textit{failed} & 5.00 / 0.42& \textbf{2.85} / 0.39\\
\hline
\multirow{4}*{\makecell{MVSEC\cite{CPYHKU:MVSEC} }}
    &Indoor Flying 1 & 5.31 / 0.37& 1.50 / 0.13& 5.09 / 0.92& 4.00 / 0.50& \textit{failed} &  1.35 / \textbf{0.11}& \textbf{0.94} / 0.14\\
~ &Indoor Flying 2 & 5.65 / 0.41& 6.98 / 0.15& \textit{failed} & 3.66 / 0.43& \textit{failed} & 1.00 / 0.16& \textbf{1.00} / \textbf{0.11}\\
~ &Indoor Flying 3 & 2.90 / 0.30& 0.73 / 0.048& 2.58 / 1.25& 1.71 / 0.18&  \textit{failed} & 0.64 / 0.065& \textbf{0.47} / \textbf{0.043}\\
~ &Indoor Flying 4 & 6.99 / 0.79& 3.62 / 0.39& \textit{failed} &\textit{failed}  & \textbf{2.77} / \textbf{0.14}& 5.31 / 0.23& 5.55 / 0.21\\
\hline
        \end{tabular}
        % \begin{tablenotes} 
        % \item \textit{Unit: m, 0.45 means the absolute position error would be 0.45m in the sequence.} 
        % \end{tablenotes} 
        \end{threeparttable} 
        }
        \end{center}
% \end{adjustbox}
        \vspace{-2.5em}
\end{table*}
% %%%%%%%%%%%%%%%%%%%%%%%%%%%%%%%%%%%%%%%%%%%%%%%%%%%%%%%%%%%%%%%%%%%%%%%%%%%%%%%%%%%%%%%%%%%%%%%%%%%%%%%%%%%%%%%%%%%%%%%%%%%%%%%%%%%%%%%%%%%%%%%%%%%%%%%%

As can be seen in Table \ref{public_dataset}, our proposed ESVIO can achieve fairly good results in most of the sequences.
Although the MPE criterion of ORB-SLAM3 is slightly better than ours in some sequences (e.g. \textit{robot-normal}, \textit{desk-normal}, \textit{mountain-normal}), our ESVIO provides more reliable and accurate results in most of the sequences under harsh situations with HDR or aggressive motion.
The proposed ESVIO is more precise than our previous PL-EVIO, especially in large-scale environments, this might be due to the better event-corner depth estimation.
While the traditional event-based methods \cite{CPYHKU:EVO}  \cite{CPYHKU:Ultimate-SLAM} \cite{CPYHKU:ESVO} failed in most of the sequences in these two datasets.

Please note that we think that parameter tuning is infeasible. 
Therefore, we evaluate our methods using fixed parameters for all sequences during the evaluations. 
However, the generalization capability of \cite{CPYHKU:EVO} and \cite{CPYHKU:ESVO} is slightly poor. Although we have put the utmost effort to tune parameters, they fail in most sequences.
Besides, we emphasize real-time performance when evaluating our methods.
Table \ref{Computing_time} illustrates the running time of our modules under different resolutions.
We also provide a qualitative comparison between our method and the other methods in the accompanying video\footnote{\url{https://b23.tv/JTkDvCP}} and supplementary material.

Last but not the least, although our proposed ESVIO achieves satisfactory results, it still has limitations in the low-texture environment.
For example, the scenarios in sequence \textit{units-dolly} and \textit{units-scooter} are so special that the visual-only method might be easy to degenerate or mismatch during the loop-closure detection.
This also indicates that either the event camera or the standard camera has limitations.
Although event cameras play a complementary role to the traditional image-based method, event-based multi-sensor fusion, especially combining non-vision-based sensors (such as GPS, and lidar), should be further developed to exploit the advantage of different sensors.

%%%%%%%%%%%%%%%%%%%%%%%%%%%%%%%%%%%%%%%%%%%%%%%%%%%%%%%%%%%%%%%%%%%%%%%%%%%%%%%%
\begin{table}[htbp]
        \vspace{-0.5em}%调整表格与正文的距离
% \begin{adjustbox}{minipage=\linewidth,bgcolor=yellow}
        \setlength{\abovecaptionskip}{-0.02cm}%调整图片标题与图距离
        \renewcommand\arraystretch{1.2}
        \tiny %此处写字体大小控制命令
        % \linespread{1.6}
        \begin{center}
        \caption{Running time of our modules in different resolution event cameras (ms)}
        \label{Computing_time}
        \resizebox{\columnwidth*1}{!}
        { 
        \begin{threeparttable}
        \begin{tabular}{ccc} 
%tabular环境是LaTeX默认创建表格的环境。你需要对这个环境手动指定一个参数。{c c c}参数告诉LaTeX，表格将会有三列，每一列都是居中对齐（c: center）
        \hline  %这个参数会在表格中插入水平的分割线。你可以多次使用这个命令。
        Modules & 346 $\times$ 260 & 640 $\times$ 480\\
        \hline
        Motion compensation & 2.70 &11.11 \\
        % Creation of SAE & 3.74 & 18.93 \\
        Creation of event representation & 5.12 & 15.47 \\
        Spatial event association & 0.82 & 2.57\\
        Temporal event association & 0.83 & 3.31\\
        The whole process of front-end & 10.44 & 35.69\\
        Back-end optimization & 19.30 & 35.59 \\
        \hline        
        \end{tabular}
        \end{threeparttable} 
        }
        \end{center}
% \end{adjustbox}
        \vspace{-1.5em}
\end{table}
%%%%%%%%%%%%%%%%%%%%%%%%%%%%%%%%%%%%%%%%%%%%%%%%%%%%%%%%%%%%%%%%%%%%%%%%%%%%%%%%%%%%%%%%%%%%%%%%%%%%%%%%%%%%

%%%%%%%%%%%%%%%%%%%%%%%%%%%%%%%%%%%%%%%%%%%%%%%%%%%%%%%%%%%%%%%%%%%%%%%%%%%%%%%%
\vspace{-1.0em}
\subsection {Indoor Quadrotor Flight Evaluation}
\label{Indoor Quadrotor Flighting Evaluation}

To further demonstrate the practicability of our methods, we perform real-world experiments on a self-designed quadrotor platform.
We choose Pixracer autopilot as our flight platform with T-Motor F90 PRO, as shown in Fig. \ref{quadrotor_flight}. 
% Our algorithm runs real-time for onboard quadrotor feed-back control.
The states estimate from our ESVIO is used to provide onboard pose feedback control for the quadrotor.
The quadrotor is commanded to follow a circular pattern eight times continuously during the experiment.
The robust and accurate onboard state estimates of our ESVIO enable real-time feedback control.
Meanwhile, we also record the ground truth from VICON for further quantitative evaluation.
% A series of real-drone experiments are conducted as follows:

%%%%%%%%%%%%i%%%%%***********************************************************************************************************************************************************%%%%%%%%%%%%%%%%%%
\begin{figure}[htb]  %%(h 此处（here） t 页顶（top）b 页底（bottom） p 独立一页（page）)
    % \vspace{-1.0em}%调整表格与正文的距离
% \begin{adjustbox}{minipage=\linewidth,bgcolor=yellow}
    % \setlength{\abovecaptionskip}{-0.1em}%调整图片标题与图距离
    	% \subfigtopskip=0pt %设置子图与上面正文或别的内容的距离
    \subfigbottomskip=-5pt %设置第二行子图与第一行子图的距离，即下面的头与上面的脚的距离
    \subfigcapskip=-10pt %设置子图与子标题之间的距离
    \captionsetup{justification=justified}%图题对齐
    \centering
    \subfigure[]{
            \begin{minipage}[t]{0.95\columnwidth}
            \centering
            \includegraphics[width=1.0\columnwidth]{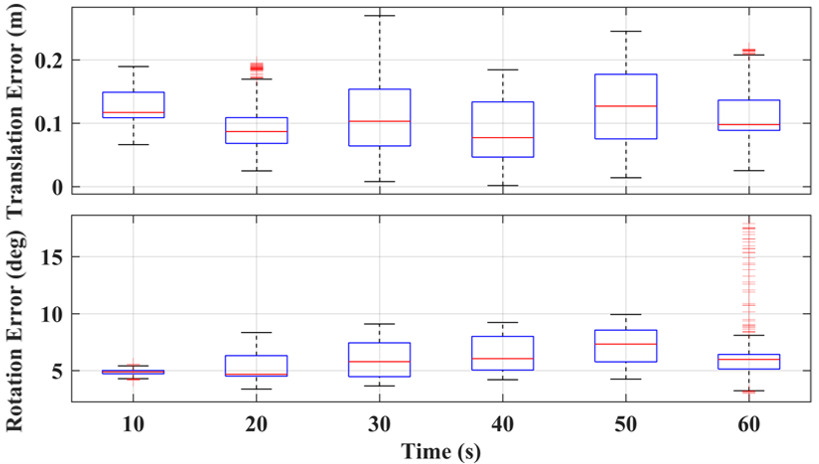}
            \label{flighting_hdr}
            \end{minipage}%
    }%
    \hspace{-5mm}
    \subfigure[]{
            \begin{minipage}[t]{0.95\columnwidth}
            \centering
            \includegraphics[width=1.0\columnwidth]{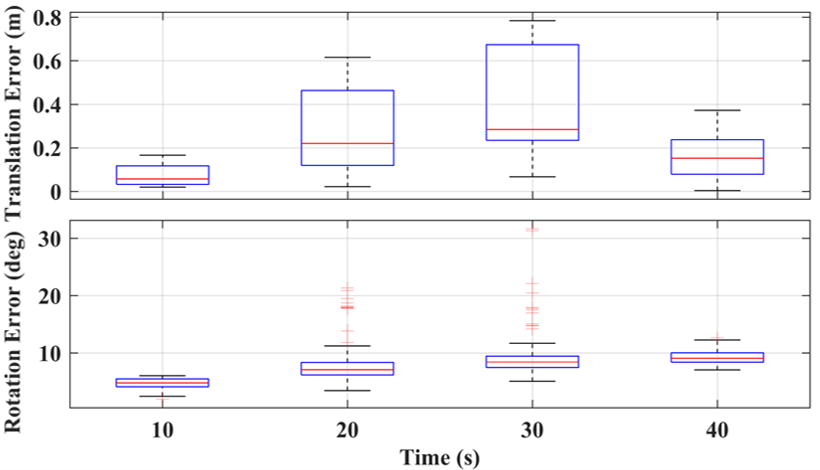}
            \label{flighting_aggressive}
            \end{minipage}%
    }%
    \caption{
    The relative error comparison of our proposed ESVIO with the VICON:
    (a) Onboard quadrotor flight in low-illumination conditions;
    (b) Onboard quadrotor flight in aggressive motion.
    }%设置图片的一个编号以及为图片添加标题
    \label{flighting_performance}
% \end{adjustbox}
    \vspace{-1.5em}%调整表格与正文的距离
\end{figure}%
%%%%%%%%%%%%%%%%%%***********************************************************************************************************************************************************%%%%%%%%%%%%%%%%%%

%%%%%%%%%%%%%%%%%%%%%%%%%%%%%%%%%%%%%%%%%%%%%%%%%%%%%%%%%%%%%%%%%%%%%%%%%%%%%%%%
\subsubsection[Section Title]{\bf Quadrotor Flight in HDR Scenarios \footnote{\url{https://b23.tv/wcZiKzG}}}
We show the relative pose error (RPE) of our ESVIO against the VICON in Fig. \ref{flighting_hdr}. 
The total trajectory length is 56.0m. 
The boxplot \cite{CPYHKU:ETH_boxplot} shows that the average relative error for the translation part is around 0.1m. 
For the rotational part, the average relative error is around $7^{\circ}$.
While there are many outliers from 50 to 60 seconds, which is caused by rapid change in yaw at that moment resulting in the estimated pose being slightly slower than VICON. 
The root-mean-square error (RMSE) of absolute trajectory error in HDR flight is 0.17m. 
Fig. \ref{firstpic} qualitatively evaluates the moment when the yaw angle changes rapidly, where few features can be extracted and tracked by the image thread, while event-corner features can still be tracked well.
% The video demo can be seen in\footnote{\url{https://b23.tv/wcZiKzG}}.

%%%%%%%%%%%%%%%%%%%%%%%%%%%%%%%%%%%%%%%%%%%%%%%%%%%%%%%%%%%%%%%%%%%%%%%%%%%%%%%%
\subsubsection[Section Title]{\bf Quadrotor Flight in Aggressive Motion \footnote{\url{https://b23.tv/nQUbMjy}}}
In this section, the yaw angle of the commanded pattern is changed drastically, for aggressive motion.
The performance of our ESVIO is qualitatively demonstrated in Fig. \ref{firstpic} and quantitatively evaluated in Fig.\ref{flighting_aggressive}. 
The RMSE of ATE in aggressive flight is 0.26m. 
Note that it would have some outliers during the comparison with the VICON. 
For example, there are some rotation errors of more than $20^\circ$ within 20-30 seconds. 
This is caused by VICON's ball is not well observed during the aggressive flight, resulting in an inaccurate measurement of the VICON at that moment.
However, our reliable ESVIO state estimator still provides robust and accurate onboard pose feedback for the quadrotor.
% The video demo can be seen in\footnote{\url{https://b23.tv/nQUbMjy}}.

%%%%%%%%%%%%%%%%%%%%%%%%%%%%%%%%%%%%%%%%%%%%%%%%%%%%%%%%%%%%%%%%%%%%%%%%%%%%%%%%
\subsection {Outdoor Large-scale Evaluation}
\label{Outdoor Large-scale Evaluation}
In this section, we evaluate our ESIO and ESVIO in outdoor large-scale environments, including the public-available autonomous driving dataset and the self-collected HKU campus dataset (More details can be seen on our website).

\subsubsection[Section Title]{\bf DSEC Dataset}
\label{DSEC}
\begin{figure}[htb]  %%(h 此处（here） t 页顶（top）b 页底（bottom） p 独立一页（page）)
        \vspace{-1.0em}%调整表格与正文的距离
% \begin{adjustbox}{minipage=\linewidth,bgcolor=yellow}
        \centering
        \captionsetup{justification=justified}%图题对齐
        \includegraphics[width=1.0\columnwidth]{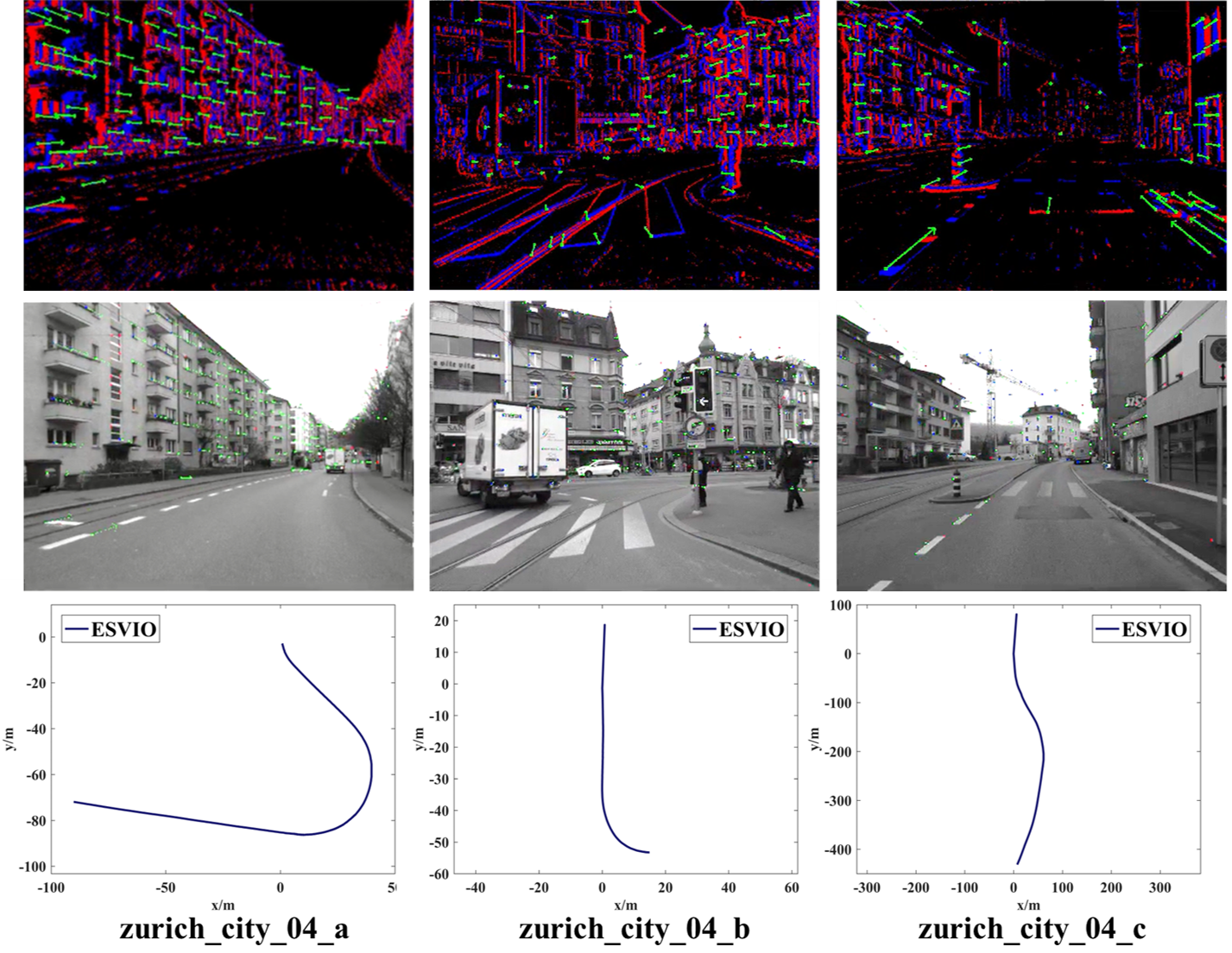}
        \caption{The qualitative results of ESVIO in DSEC dataset for sequences zurich\_city\_04 (a) to (c). \textbf{Top:} The stereo event-corner feature tracking performance; \textbf{Middle:} The stereo image-based feature tracking performance; \textbf{Bottom:} The estimated trajectories produced by our ESVIO.
        }  %设置图片的一个编号以及为图片添加标题
        \label{DSEC_ESVIO_part}
% \end{adjustbox}
        \vspace{-0.5em}%调整表格与正文的距离
\end{figure}

DSEC \cite{CPYHKU:DSEC} is collected by high-resolution stereo event cameras (640 $\times$ 480) under driving scenarios, which is challenging for event-based sensors, as forward motions typically produce considerably fewer events at the center. 
Qualitative evaluation can be seen in Fig. \ref{DSEC_ESVIO_part}. 
Since the DSEC dataset does not provide the ground truth 6-DoF poses, we only show the estimated trajectory and the tracking performance of our event-based and image-based features.
Both our ESIO (available in supplementary material) and ESVIO can achieve satisfactory results.

%%%%%%%%%%%%%%%%%%***********************************************************************************************************************************************************%%%%%%%%%%%%%%%%%%
%%%%%%%%%%%%%%%%%%***********************************************************************************************************************************************************%%%%%%%%%%%%%%%%%%

%%%%%%%%%%%%%%%%%%%%%%%%%%%%%%%%%%%%%%%%%%%%%%%%%%%%%%%%%%%%%%%%%%%%%%%%%%%%%%%%

\subsubsection{\bf HKU Large-scale environment}
\label{outdoor large-scale}
\begin{figure}[htb]  %%(h 此处（here） t 页顶（top）b 页底（bottom） p 独立一页（page）)
        % \setlength{\abovecaptionskip}{-1.0em}%调整图片标题与图距离
        % \vspace{-1.0em}%调整表格与正文的距离
        \centering
        \captionsetup{justification=justified}%图题对齐
        \includegraphics[width=1.0\columnwidth]{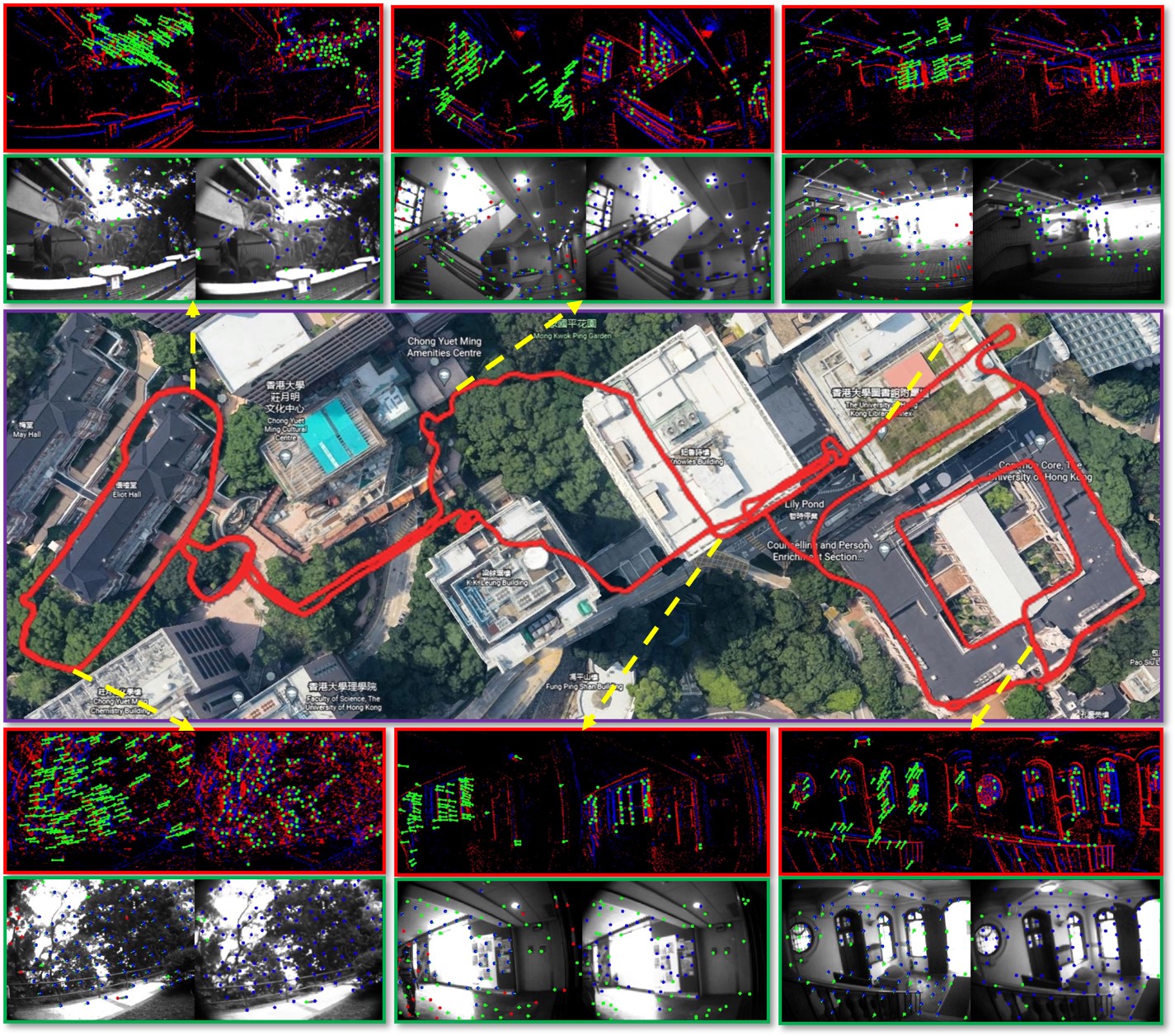}
        \caption{The estimated trajectory of our ESVIO in the large-scale environment. 
        The detection and tracking situation of the stereo event-corner features and stereo image features during the experiment are also visualized.
        }  %设置图片的一个编号以及为图片添加标题
        \label{google_map}
        \vspace{-1.0em}%调整表格与正文的距离
\end{figure}

This section carried out a large-scale experiment on the HKU campus to illustrate the long-time practicability of our ESVIO, features with large-scale, indoor-outdoor conversion, pedestrians in the scene generating outlier events, etc. 
The path length of the outdoor evaluation is around 1.8 km and the duration is 34.9 minutes. 
The evaluation covers the place around 310m in length, 170m in width, and 55m in height changes.
Since the VICON is not available outdoors, we only show the qualitative performance and the estimated trajectory overlaid with the Google map for visual comparison.
As can be seen from Fig. \ref{google_map}, our ESVIO performs well in long-term motion evaluation, the estimated trajectory is aligned and almost coincide with the Google map.
% More details of this outdoor large-scale evaluation can be seen on our website.

%%%%%%%%%%%%%%%%%%***********************************************************************************************************************************************************%%%%%%%%%%%%%%%%%%
%%%%%%%%%%%%%%%%%%***********************************************************************************************************************************************************%%%%%%%%%%%%%%%%%%

\section{CONCLUSIONS}
\label{CONCLUSIONS}
In this paper, we propose a robust, real-time event-based stereo VIO, ESVIO, which tightly fuses the stereo event streams, stereo image frames, and IMU measurements using sliding windows graph-based optimization. 
Geometry-based spatial and temporal associations between consecutive stereo event streams are designed to ensure robust state estimation.
In addition, the motion compensation approach corrects the curved event streams with IMU and ESVIO back-end to emphasize the contour of scenes.
Extensive evaluations demonstrate that our ESVIO achieves superior performance compared to other state-of-the-art algorithms on public datasets and our self-collected challenging datasets.
Furthermore, we also perform various onboard closed-loop flights using the proposed ESVIO under low-light scenes and aggressive motion.
In our future work, we might further explore the event-based mapping for the quadrotor system which should be able to support autonomous navigation and obstacle avoidance.

%%%%%%%%%%%%%%%%%%%%%%%%%%%%%%%%%%%%%%%%%%%%%%%%%%%%%%%%%%%%%%%%%%%%%%%%%%%%%%%%
\bibliographystyle{IEEEtran} % 参考文献排版风格，这个是IEEE transaction的，其他可以自查
\bibliography{references_peiyu.bib} % 导入bib，references为“references.bib"的文件名

\begin{thebibliography}{10}
\providecommand{\url}[1]{#1}
\csname url@rmstyle\endcsname
\providecommand{\newblock}{\relax}
\providecommand{\bibinfo}[2]{#2}
\providecommand\BIBentrySTDinterwordspacing{\spaceskip=0pt\relax}
\providecommand\BIBentryALTinterwordstretchfactor{4}
\providecommand\BIBentryALTinterwordspacing{\spaceskip=\fontdimen2\font plus
\BIBentryALTinterwordstretchfactor\fontdimen3\font minus
  \fontdimen4\font\relax}
\providecommand\BIBforeignlanguage[2]{{%
\expandafter\ifx\csname l@#1\endcsname\relax
\typeout{** WARNING: IEEEtran.bst: No hyphenation pattern has been}%
\typeout{** loaded for the language `#1'. Using the pattern for}%
\typeout{** the default language instead.}%
\else
\language=\csname l@#1\endcsname
\fi
#2}}

\bibitem{CPYHKU:EVENT-SURVEY}
G.~Gallego, T.~Delbr{\"u}ck, G.~Orchard, C.~Bartolozzi, B.~Taba, A.~Censi,
  S.~Leutenegger, A.~J. Davison, J.~Conradt, K.~Daniilidis, \emph{et~al.},
  ``Event-based vision: A survey,'' \emph{IEEE transactions on pattern analysis
  and machine intelligence}, vol.~44, no.~1, pp. 154--180, 2020.

\bibitem{CPYHKU:EVO}
H.~Rebecq, T.~Horstsch{\"a}fer, G.~Gallego, and D.~Scaramuzza, ``Evo: A
  geometric approach to event-based 6-dof parallel tracking and mapping in real
  time,'' \emph{IEEE Robotics and Automation Letters}, vol.~2, no.~2, pp.
  593--600, 2017.

\bibitem{CPYHKU:Ultimate-SLAM}
A.~R. Vidal, H.~Rebecq, T.~Horstschaefer, and D.~Scaramuzza, ``Ultimate slam?
  combining events, images, and imu for robust visual slam in hdr and
  high-speed scenarios,'' \emph{IEEE Robotics and Automation Letters}, vol.~3,
  no.~2, pp. 994--1001, 2018.

\bibitem{CPYHKU:PL-EVIO}
W.~Guan, P.~Chen, Y.~Xie, and P.~Lu, ``Pl-evio: Robust monocular event-based
  visual inertial odometry with point and line features,'' \emph{arXiv preprint
  arXiv:2209.12160}, 2022.

\bibitem{CPYHKU:ESVO}
Y.~Zhou, G.~Gallego, and S.~Shen, ``Event-based stereo visual odometry,''
  \emph{IEEE Transactions on Robotics}, 2021.

\bibitem{CPYHKU:Feature-based-ESVO}
A.~Hadviger, I.~Cvi{\v{s}}i{\'c}, I.~Markovi{\'c}, S.~Vra{\v{z}}i{\'c}, and
  I.~Petrovi{\'c}, ``Feature-based event stereo visual odometry,'' in
  \emph{2021 European Conference on Mobile Robots (ECMR)}.\hskip 1em plus 0.5em
  minus 0.4em\relax IEEE, 2021, pp. 1--6.

\bibitem{CPYHKU:kueng2016low}
B.~Kueng, E.~Mueggler, G.~Gallego, and D.~Scaramuzza, ``Low-latency visual
  odometry using event-based feature tracks,'' in \emph{2016 IEEE/RSJ
  International Conference on Intelligent Robots and Systems (IROS)}.\hskip 1em
  plus 0.5em minus 0.4em\relax IEEE, 2016, pp. 16--23.

\bibitem{CPYHKU:EMVS}
H.~Rebecq, G.~Gallego, E.~Mueggler, and D.~Scaramuzza, ``Emvs: Event-based
  multi-view stereo—3d reconstruction with an event camera in real-time,''
  \emph{International Journal of Computer Vision}, vol. 126, no.~12, pp.
  1394--1414, 2018.

\bibitem{CPYHKU:Event-based-visual-inertial-odometry}
A.~Zihao~Zhu, N.~Atanasov, and K.~Daniilidis, ``Event-based visual inertial
  odometry,'' in \emph{Proceedings of the IEEE Conference on Computer Vision
  and Pattern Recognition}, 2017, pp. 5391--5399.

\bibitem{CPYHKU:ETH-EIO}
H.~Rebecq, T.~Horstschaefer, and D.~Scaramuzza, ``Real-time visual-inertial
  odometry for event cameras using keyframe-based nonlinear optimization,'' in
  \emph{British Machine Vision Conference (BMVC)}, 2017.

\bibitem{CPYHKU:Continuous-time-visual-inertial-odometry-for-event-cameras}
E.~Mueggler, G.~Gallego, H.~Rebecq, and D.~Scaramuzza, ``Continuous-time
  visual-inertial odometry for event cameras,'' \emph{IEEE Transactions on
  Robotics}, vol.~34, no.~6, pp. 1425--1440, 2018.

\bibitem{CPYHKU:DEVO}
Y.~Zuo, J.~Yang, J.~Chen, X.~Wang, Y.~Wang, and L.~Kneip, ``Devo: depth-event
  camera visual odometry in challenging conditions,'' in \emph{2022
  International Conference on Robotics and Automation (ICRA)}.\hskip 1em plus
  0.5em minus 0.4em\relax IEEE, 2022, pp. 2179--2185.

\bibitem{CPYHKU:EKLT-VIO}
F.~Mahlknecht, D.~Gehrig, J.~Nash, F.~M. Rockenbauer, B.~Morrell, J.~Delaune,
  and D.~Scaramuzza, ``Exploring event camera-based odometry for planetary
  robots,'' \emph{IEEE Robotics and Automation Letters (RA-L)}, 2022.

\bibitem{CPYHKU:EKLT}
D.~Gehrig, H.~Rebecq, G.~Gallego, and D.~Scaramuzza, ``Eklt: Asynchronous
  photometric feature tracking using events and frames,'' \emph{International
  Journal of Computer Vision}, vol. 128, no.~3, pp. 601--618, 2020.

\bibitem{CPYHKU:GuanEVIO}
W.~Guan and P.~Lu, ``Monocular event visual inertial odometry based on
  event-corner using sliding windows graph-based optimization,'' in \emph{2022
  IEEE/RSJ International Conference on Intelligent Robots and Systems
  (IROS)}.\hskip 1em plus 0.5em minus 0.4em\relax IEEE, 2022, pp. 2438--2445.

\bibitem{tulyakov2019learning}
S.~Tulyakov, F.~Fleuret, M.~Kiefel, P.~Gehler, and M.~Hirsch, ``Learning an
  event sequence embedding for dense event-based deep stereo,'' in
  \emph{Proceedings of the IEEE/CVF International Conference on Computer
  Vision}, 2019, pp. 1527--1537.

\bibitem{nam2022stereo}
Y.~Nam, M.~Mostafavi, K.-J. Yoon, and J.~Choi, ``Stereo depth from events
  cameras: Concentrate and focus on the future,'' in \emph{Proceedings of the
  IEEE/CVF Conference on Computer Vision and Pattern Recognition}, 2022, pp.
  6114--6123.

\bibitem{CPYHKU:ETH_SCI_Ro_avo}
D.~Falanga, K.~Kleber, and D.~Scaramuzza, ``Dynamic obstacle avoidance for
  quadrotors with event cameras,'' \emph{Science Robotics}, vol.~5, no.~40, p.
  eaaz9712, 2020.

\bibitem{CPYHKU:FAST-Dynamic-Vision}
B.~He, H.~Li, S.~Wu, D.~Wang, Z.~Zhang, Q.~Dong, C.~Xu, and F.~Gao,
  ``Fast-dynamic-vision: Detection and tracking dynamic objects with event and
  depth sensing,'' in \emph{2021 IEEE/RSJ International Conference on
  Intelligent Robots and Systems (IROS)}, 2021, pp. 3071--3078.

\bibitem{CPYHKU:LK_optical_flow}
B.~D. Lucas, T.~Kanade, \emph{et~al.}, ``An iterative image registration
  technique with an application to stereo vision.''\hskip 1em plus 0.5em minus
  0.4em\relax Vancouver, British Columbia, 1981.

\bibitem{CPYHKU:ARC*}
I.~Alzugaray and M.~Chli, ``Asynchronous corner detection and tracking for
  event cameras in real time,'' \emph{IEEE Robotics and Automation Letters},
  vol.~3, no.~4, pp. 3177--3184, 2018.

\bibitem{CPYHKU:VINS-Fusion}
T.~Qin, J.~Pan, S.~Cao, and S.~Shen, ``A general optimization-based framework
  for local odometry estimation with multiple sensors,'' \emph{arXiv preprint
  arXiv:1901.03638}, 2019.

\bibitem{CPYHKU:MVSEC}
A.~Z. Zhu, D.~Thakur, T.~{\"O}zaslan, B.~Pfrommer, V.~Kumar, and K.~Daniilidis,
  ``The multivehicle stereo event camera dataset: An event camera dataset for
  3d perception,'' \emph{IEEE Robotics and Automation Letters}, vol.~3, no.~3,
  pp. 2032--2039, 2018.

\bibitem{CPYHKU:VECtor}
L.~Gao, Y.~Liang, J.~Yang, S.~Wu, C.~Wang, J.~Chen, and L.~Kneip, ``Vector: A
  versatile event-centric benchmark for multi-sensor slam,'' \emph{IEEE
  Robotics and Automation Letters}, 2022.

\bibitem{CPYHKU:evo_package}
M.~Grupp, ``evo: Python package for the evaluation of odometry and slam,''
  \emph{Note: https://github. com/MichaelGrupp/evo Cited by: Table}, vol.~7,
  2017.

\bibitem{CPYHKU:ORB-SLAM3}
C.~Campos, R.~Elvira, J.~J.~G. Rodr{\'\i}guez, J.~M. Montiel, and J.~D.
  Tard{\'o}s, ``Orb-slam3: An accurate open-source library for visual,
  visual--inertial, and multimap slam,'' \emph{IEEE Transactions on Robotics},
  2021.

\bibitem{CPYHKU:ETH_boxplot}
Z.~Zhang and D.~Scaramuzza, ``A tutorial on quantitative trajectory evaluation
  for visual(-inertial) odometry,'' in \emph{2018 IEEE/RSJ International
  Conference on Intelligent Robots and Systems (IROS)}.\hskip 1em plus 0.5em
  minus 0.4em\relax IEEE, 2018, pp. 7244--7251.

\bibitem{CPYHKU:DSEC}
M.~Gehrig, W.~Aarents, D.~Gehrig, and D.~Scaramuzza, ``Dsec: A stereo event
  camera dataset for driving scenarios,'' \emph{IEEE Robotics and Automation
  Letters}, vol.~6, no.~3, pp. 4947--4954, 2021.

\end{thebibliography}

\includepdf[pages=-]{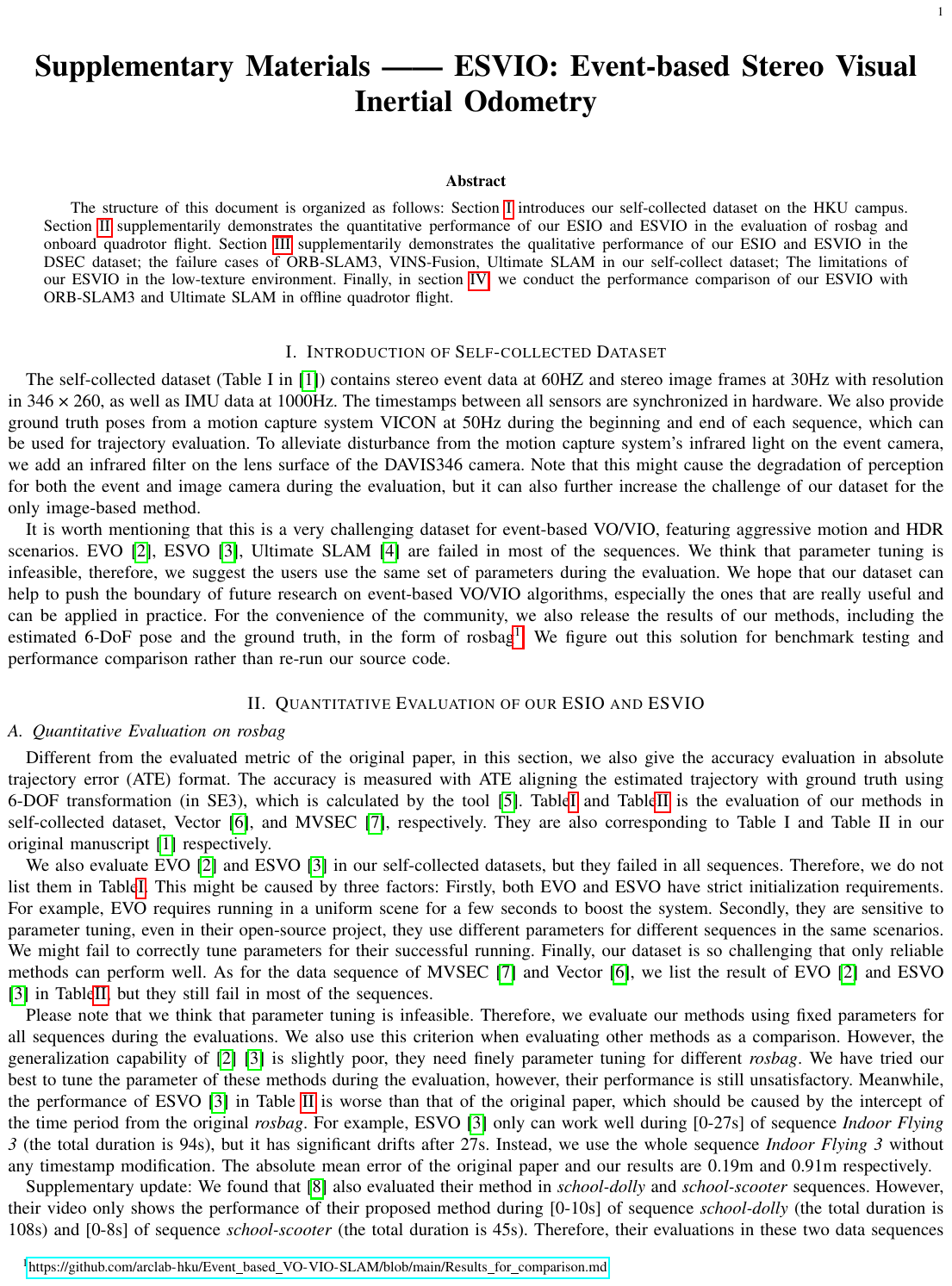}

\end{document}